\documentclass[lettersize,journal]{IEEEtran}
\usepackage{amsmath,amsfonts}
\usepackage{algorithmic}
\usepackage{algorithm}
\usepackage{array}
\usepackage[caption=false,font=normalsize,labelfont=sf,textfont=sf]{subfig}
\usepackage{textcomp}
\usepackage{stfloats}
\usepackage{url}
\usepackage{verbatim}
\usepackage{graphicx}
\usepackage{cite}
\usepackage{hyperref}
\usepackage{multirow}
\usepackage{wrapfig}
\usepackage{xspace}
\usepackage{amsmath}
\usepackage[capitalise,noabbrev,nameinlink]{cleveref}
\crefformat{equation}{#2Equation~#1#3}
\usepackage{pgfplots,pgfplotstable} 
\usepackage{subcaption}
\usepackage{color}
\usepackage{colortbl}
\usepackage{marvosym}
\usepackage{booktabs}
\usepackage{stfloats}
\usepackage{ulem}

\usepgfplotslibrary{colorbrewer,groupplots}
\usetikzlibrary{positioning}
\pgfplotsset{compat=1.18}
\newcommand{\eg}{e.g.\@\xspace}
\newcommand{\ie}{i.e.\@\xspace}

\definecolor{darklavender}{rgb}{0.45, 0.31, 0.59}
\definecolor{darkviolet}{rgb}{0.58, 0.0, 0.83}
\definecolor{Coral}{rgb}{1, 0.47, 0.24}

\definecolor{visible-blue}{rgb}{0.286, 0.525, 0.910}
\definecolor{tabfirst}{rgb}{1, 0.7, 0.7} 
\definecolor{tabsecond}{rgb}{1, 0.85, 0.7} 
\definecolor{tabthird}{rgb}{1, 1, 0.7} 

\newcommand{\modelname}{Octree-GS\xspace}

\newcommand{\OOM}[1]{-}

\definecolor{placeholder}{rgb}{0.6,0.8,0.95}

\begin{document}

\title{\modelname: Towards Consistent Real-time Rendering with LOD-Structured 3D Gaussians}

\author{
Kerui Ren$^*$, 
Lihan Jiang$^*$,
Tao Lu,
Mulin Yu,
Linning Xu,
Zhangkai Ni,
Bo Dai$^\dag$
\thanks{K. Ren is with Shanghai Jiao Tong University and Shanghai AI Laboratory. E-mail: renkerui@sjtu.edu.cn.}
\thanks{L. Jiang is with The University of Science and Technology of China and Shanghai AI Laboratory. E-mail: jianglihan@mail.ustc.edu.cn.}
\thanks{T. Lu is with Brown University. E-mail: tao\_lu@brown.edu.}
\thanks{B. Dai and M. Yu are with Shanghai AI Laboratory. E-mails: doubledaibo@gmail.com, yumulin@pjlab.org.cn.}
\thanks{L. Xu is with The Chinese University of Hong Kong. E-mail: linningxu@link.cuhk.edu.hk.}
\thanks{Z. Ni is with Tongji University.}
\thanks{$*$ Equal contribution.}
\thanks{$\dag$ Corresponding author.}
}

\markboth{IEEE TRANSACTIONS ON PATTERN ANALYSIS AND MACHINE INTELLIGENCE}%
{How to Use the IEEEtran \LaTeX \ Templates}


\maketitle
\begin{figure*}[t!]
\includegraphics[width=\linewidth]{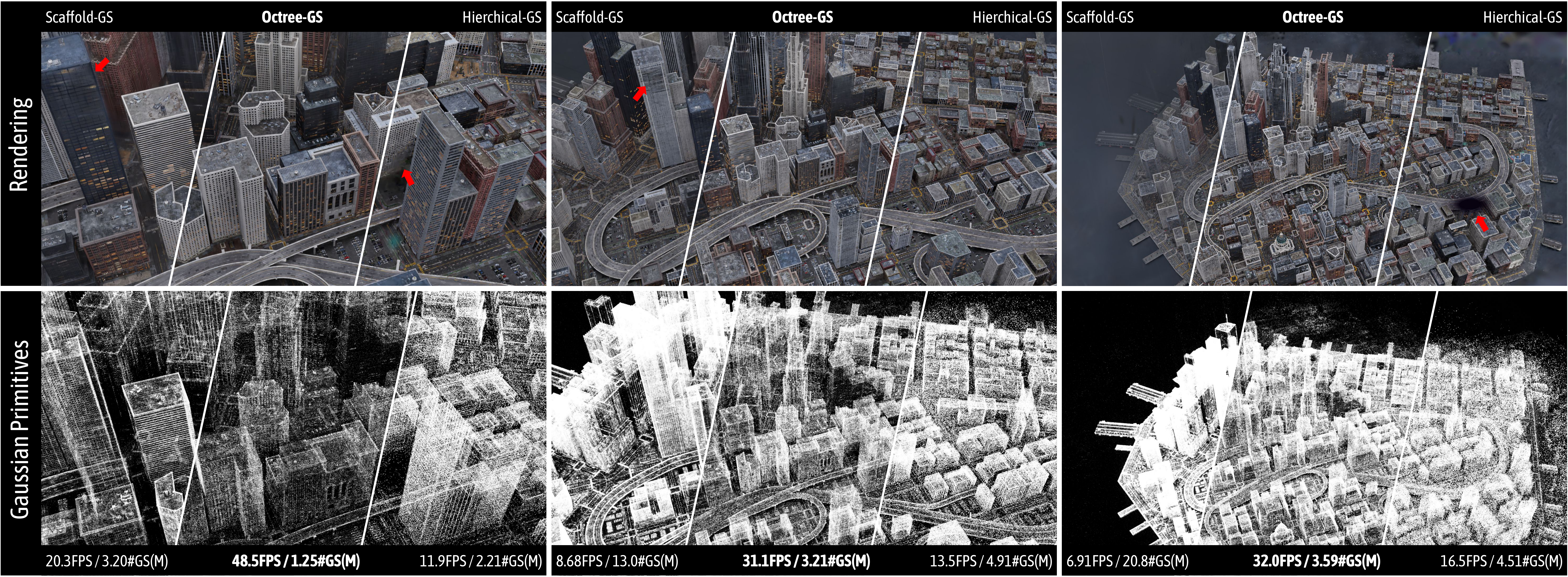}
\caption{
Visualization of a continuous zoom-out trajectory on the MatrixCity~\cite{li2023matrixcity} dataset.
Both the rendered 2D images and the corresponding Gaussian primitives are indicated.
As indicated by the highlighted arrows, \modelname consistently demonstrates superior visual quality compared to state-of-the-art methods Hierarchical-GS~\cite{kerbl2024hierarchical} and Scaffold-GS~\cite{lu2023scaffold}. Both SOTA methods fail to render the excessive number of Gaussian primitives included in distant views in real-time, whereas \modelname consistently achieves real-time rendering performance ($\geq 30$ FPS). First row metrics: FPS/storage size.
}
\vspace{-1em}
\label{fig:teaser}
\end{figure*}

\begin{abstract}

The recently proposed 3D Gaussian Splatting (3D-GS) demonstrates superior rendering fidelity and efficiency compared to NeRF-based scene representations.
However, it struggles in large-scale scenes due to the high number of Gaussian primitives, particularly in zoomed-out views, where all primitives are rendered regardless of their projected size. This often results in inefficient use of model capacity and difficulty capturing details at varying scales.
To address this, we introduce Octree-GS, a Level-of-Detail (LOD) structured approach that dynamically selects appropriate levels from a set of multi-scale Gaussian primitives, ensuring consistent rendering performance. 
To adapt the design of LOD, we employ an innovative grow-and-prune strategy for densification and also propose a progressive training strategy to arrange Gaussians into appropriate LOD levels.
Additionally, our LOD strategy generalizes to other Gaussian-based methods, such as 2D-GS and Scaffold-GS, reducing the number of primitives needed for rendering while maintaining scene reconstruction accuracy. 
Experiments on diverse datasets demonstrate that our method achieves real-time speeds, with even 10 $\times$ faster than state-of-the-art methods in large-scale scenes, without compromising visual quality. Project page: \href{https://city-super.github.io/octree-gs/}{\textcolor{magenta}{\textbf{https://city-super.github.io/octree-gs/}}}.

\end{abstract}

\begin{IEEEkeywords}
Novel View Synthesis, 3D Gaussian Splatting, Consistent Real-time Rendering, Level-of-Detail
\end{IEEEkeywords}

\section{Introduction}
\label{sec:intro}
\IEEEPARstart{T}{he} 
field of novel view synthesis has seen significant advancements driven by the advancement of radiance fields~\cite{mildenhall2021nerf}, which deliver high-fidelity rendering. However, these methods often suffer from slow training and rendering speeds due to time-consuming stochastic sampling. Recently, 3D Gaussian splatting (3D-GS)~\cite{kerbl20233d} has pushed the field forward by using anisotropic Gaussian primitives, achieving near-perfect visual quality with efficient training times and tile-based splatting techniques for real-time rendering.
With such strengths, it has significantly accelerated the process of replicating the real world into a digital counterpart~\cite{zielonka2023drivable, saito2023relightable, zheng2023gps, qian2023gaussianavatars}, igniting the community's imagination for scaling real-to-simulation environments~\cite{yan2024street, zhou2023drivinggaussian, lu2023scaffold}. With its exceptional visual effects, an unprecedented photorealistic experience in VR/AR~\cite{jiang2024vr, xie2023physgaussian} is now more attainable than ever before.

A key drawback of 3D-GS~\cite{kerbl20233d} is the misalignment between the distribution of 3D Gaussians and the actual scene structure. Instead of aligning with the geometry of the scene, the Gaussian primitives are distributed based on their fit to the training views, leading to inaccurate and inefficient placement. This misalignment causes two bottleneck challenges: 1) it reduces robustness in rendering views that differ significantly from the training set, as the primitives are not optimized for generalization, and 2) results in redundant and overlap primitives that fail to efficiently represent scene details for real-time rendering, especially in large-scale urban scenes with millions of primitives.

There are variants of the vanilla 3D-GS~\cite{kerbl20233d} that aim at resolving the misalignment between the organization of 3D Gaussians and the structure of target scene.
Scaffold-GS~\cite{lu2023scaffold} enhances the structure alignment by introducing a regularly spaced feature grid as a structural prior, improving the arrangement and viewpoint-aware adjustment of Gaussians for better rendering quality and efficiency. Mip-Splatting~\cite{yu2023mip} resorts to 3D smoothing and 2D Mip filters to alleviate the redundancy of 3D Gaussians during the optimiziation process of 3D-GS. 2D-GS~\cite{huang20242d} forces the primitives to better align with the surface, enabling faster reconstruction.

Although the aforementioned improvements have been extensively tested on diverse public datasets, we identify a new challenge in the Gaussian era: recording large-scale scenes is becoming increasingly common, yet these methods inherently struggles to scale, as shown in Fig~\ref{fig:teaser}.
This limitation arises because they still rely on visibility-based filtering for primitive selection, considering all primitives within the view frustum without accounting for their projected sizes. As a result, every object detail is rendered, regardless of distance, leading to redundant computations and inconsistent rendering speeds, particularly in zoom-out scenarios involving large, complex scenes. The lack of Level-of-Detail (LOD) adaptation further forces all 3D Gaussians to compete across views, degrading rendering quality at different scales. As scene complexity increases, the growing number of Gaussians amplifies bottlenecks in real-time rendering.

To address the aforementioned issues and better accommodate the new era, we integrate an octree structure into the Gaussian representation, inspired by previous works~\cite{xu2023vr, yu2021plenoctrees, martel2021acorn} that demonstrate the effectiveness of spatial structures like octrees and multi-resolution grids for flexible content allocation and real-time rendering. 
Specifically,
our method organizes scenes with hierarchical grids to meet LOD needs, efficiently adapting to complex or large-scale scenes during both training and inference, with LOD levels selected based on observation footprint and scene detail richness. 
We further employ a progressive training strategy, introducing a novel growing and pruning approach. A next-level growth operator enhances connections between LODs, increasing high-frequency detail, while redundant Gaussians are pruned based on opacity and view frequency.
By adaptively querying LOD levels from the octree-based Gaussian structure based on viewing distance and scene complexity, our method minimizes the number of primitives needed for rendering, ensuring consistent efficiency, as shown in Fig.~\ref{fig:teaser}. In addition, \modelname effectively separates coarse and fine scene details, allowing for accurate Gaussian placement at appropriate scales, significantly improving reconstruction fidelity and texture detail.

Unlike other concurrent LOD methods~\cite{kerbl2024hierarchical,liu2024citygaussian}, our approach is an end-to-end algorithm that achieves LOD effects in a single training round, reducing training time and storage overhead.
Notably, our LOD framework is also compatible with various Gaussian representations, including explicit Gaussians~\cite{huang20242d,kerbl20233d} and neural Gaussians~\cite{lu2023scaffold}. By incorporating our strategy, we have demonstrated significant enhancements in visual performance and rendering speed across a wide range of datasets, including both fine-detailed indoor scenes and large-scale urban environments.

In summary, our method offers the following key contributions:
\begin{itemize}
    \item To the best of our knowledge, \modelname is the \textbf{first approach to deal with the problem of Level-of-Detail in Gaussian representation}, enabling consistent rendering speed by dynamically adjusting the fetched LOD on-the-fly owing to our explicit octree structure design.
    \item We develop a novel \textbf{grow-and-prune strategy} optimized for LOD adaptation.
    \item We introduce a \textbf{progressive training strategy} to encourage more reliable distributions of primitives.
    \item Our LOD strategy is able to \textbf{generalize to any Gaussian-based method}.
    \item Our methods, while maintaining the superior rendering quality, achieves \textbf{state-of-the-art rendering speed}, especially in \textbf{large-scale scenes} and extreme-view sequences, as shown in Fig.~\ref{fig:teaser}.
\end{itemize}

\section{Related work}
\label{sec:related_work}

\subsection{Novel View Synthesis}
NeRF methods~\cite{mildenhall2021nerf} have revolutionized the novel view synthesis task with their photorealistic rendering and view-dependent modeling effects. By leveraging classical volume rendering equations, NeRF trains a coordinate-based MLP to encode scene geometry and radiance, mapping directly from positionally encoded spatial coordinates and viewing directions.
To ease the computational load of dense sampling process and forward through deep MLP layers, researchers have resorted to various hybrid-feature grid representations, akin to `caching' intermediate latent features for final rendering
~\cite{liu2020neural,yu2021plenoctrees,fridovich2022plenoxels,sun2022direct,chen2022tensorf,muller2022instant,xu2023grid,xiangli2023assetfield}. 
Multi-resolution hash encoding~\cite{muller2022instant} is commonly chosen as the default backbone for many recent advancements due to its versatility for enabling fast and efficient rendering, encoding scene details at various granularities~\cite{turki2024pynerf, li2023neuralangelo, reiser2024binary} and extended supports for LOD renderings~\cite{xu2023vr,barron2023zip}.

Recently, 3D-GS~\cite{kerbl20233d} has ignited a revolution in the field by employing anisotropic 3D Gaussians to represent scenes, achieving
state-of-the-art rendering quality and speed.
Subsequent studies have rapidly expanded 3D-GS into diverse downstream applications beyond static 3D reconstruction, 
sparking a surge of extended applications to 3D generative modeling~\cite{tang2023dreamgaussian,liang2023luciddreamer,tang2024lgm}, physical simulation~\cite{xie2023physgaussian,feng2024gaussian}, dynamic modeling~\cite{luiten2023dynamic,yang2023deformable,huang2023sc}, SLAMs~\cite{yugay2023gaussian,keetha2023splatam},
and autonomous driving scenes~\cite{jiang2024vr, yan2024street,zhou2023drivinggaussian}, etc.
Despite the impressive rendering quality and speed of 3D-GS, its ability to sustain stable real-time rendering with rich content is hampered by the accompanying rise in resource costs. 
This limitation hampers its practicality in speed-demanding applications, such as gaming in open-world environments and other immersive experiences, particularly for large indoor and outdoor scenes with computation-restricted devices.

\subsection{Spatial Structures for Neural Scene Representations}
Various spatial structures have been explored in previous NeRF-based representations, including dense voxel grids~\cite{liu2020neural,sun2022direct}, sparse voxel grids~\cite{yu2021plenoctrees,fridovich2022plenoxels}, point clouds \cite{xu2022point}, multiple compact low-rank tensor components \cite{chen2022tensorf,fridovich2023k,cao2023hexplane}, and multi-resolution hash tables~\cite{muller2022instant}. 
These structures primarily aim to enhance training or inference speed and optimize storage efficiency. 
Inspired by classical computer graphics techniques such as BVH~\cite{rubin19803} and SVO~\cite{laine2010efficient} which are designed to model the scene in
a sparse hierarchical structure for ray tracing acceleration.
NSVF~\cite{liu2020neural} efficiently skipping the empty voxels leveraging the neural implicit fields structured in sparse octree grids.
PlenOctree~\cite{yu2021plenoctrees} stores the appearance and density values in every leaf to enable highly efficient rendering. DOT \cite{bai2023dynamic} improves the fixed octree design in Plenoctree with hierarchical feature fusion. 
ACORN~\cite{martel2021acorn} introduces a multi-scale hybrid implicit–explicit network architecture based on octree optimization.

While vanilla 3D-GS~\cite{kerbl20233d} imposes no restrictions on the spatial distribution of all 3D Gaussians, allowing the modeling of scenes with a set of initial sparse point clouds, Scaffold-GS \cite{lu2023scaffold} introduces a hierarchical structure, facilitating more accurate and efficient scene reconstruction.
In this work, we introduce a sparse octree structure to Gaussian primitives, 
which demonstrates improved capabilities such as real-time rendering stability irrespective of trajectory changes.

\subsection{Level-of-Detail (LOD)}

LOD is widely used in computer graphics to manage the complexity of 3D scenes, balancing visual quality and computational efficiency. It is crucial in various applications, including real-time graphics, CAD models, virtual environments, and simulations. 
Geometry-based LOD involves simplifying the geometric representation of 3D models using techniques like mesh decimation; while rendering-based LOD creates the illusion of detail for distant objects presented on 2D images.
The concept of LOD finds extensive applications in geometry reconstruction\cite{verdie_tog15,fang_cvpr18,Yu_cvpr22} and neural rendering \cite{barron2021mip,barron2022mip,barron2023zip,turki2024pynerf,xu2023vr}. 
Mip-NeRF~\cite{barron2021mip} addresses aliasing artifacts by cone-casting approach approximated with Gaussians. 
BungeeNeRF~\cite{xiangli2022bungeenerf} employs residual blocks and inclusive data supervision for diverse multi-scale scene reconstruction.
To incorporate LOD into efficient grid-based NeRF approaches like instant-NGP \cite{muller2022instant}, Zip-NeRF\cite{barron2023zip} further leverages supersampling as a prefiltered feature approximation. 
VR-NeRF\cite{xu2023vr} utilizes mip-mapping hash grid for continuous LOD rendering and an immersive VR experience. PyNeRF\cite{turki2024pynerf} employs a pyramid design to adaptively capture details based on scene characteristics. 
However, GS-based LOD methods fundamentally differ from above LOD-aware NeRF methods in scene representation and LOD introduction. For instance, NeRF can compute LOD from per-pixel footprint size, whereas GS-based methods require joint LOD modeling from both the view and 3D scene level. We introduce a flexible octree structure to address LOD-aware rendering in the 3D-GS framework.

Concurrent works related to our method include LetsGo~\cite{cui2024letsgo}, CityGaussian~\cite{liu2024citygaussian}, and Hierarchical-GS~\cite{kerbl2024hierarchical}, all of which also leverage LOD for large-scale scene reconstruction. 
1) LetsGo introduces multi-resolution Gaussian models optimized jointly, focusing on garage reconstruction, but requires multi-resolution point cloud inputs, leading to higher training overhead and reliance on precise point cloud accuracy, making it more suited for lidar scanning scenarios. 
2) CityGaussian selects LOD levels based on distance intervals and fuses them for efficient large-scale rendering, but lacks robustness due to the need for manual distance threshold adjustments, and faces issues like stroboscopic effects when switching between LOD levels. 
3) Hierarchical-GS, using a tree-based hierarchy, shows promising results in street-view scenes but involves post-processing for LOD, leading to increased complexity and longer training times.
A common limitation across these methods is that each LOD level independently represents the entire scene, increasing storage demands. In contrast, \modelname employs an explicit octree structure with an accumulative LOD strategy, which significantly accelerates rendering speed while reducing storage requirements.

\section{Preliminaries}
\label{preliminaries}
In this section, we present a brief overview of the core concepts underlying 3D-GS~\cite{kerbl20233d} and Scaffold-GS~\cite{lu2023scaffold}.

\subsection{3D-GS} 
\label{pre: 3d-gs}
3D Gaussian splatting~\cite{kerbl20233d} explicitly models scenes using anisotropic 3D Gaussians and renders images by rasterizing the projected 2D counterparts.
Each 3D Gaussian $G(x)$ is parameterized by a center position $\mu \in \mathbb{R}^3$  and a covariance $\Sigma \in \mathbb{R}^{3 \times 3}$:
\begin{equation}
    G(x)=e^{-\frac{1}{2}(x-\mu)^T \Sigma^{-1}(x-\mu)},
\end{equation}
where $x$ is an arbitrary position within the scene,
$\Sigma$ is parameterized by a scaling matrix $S \in \mathbb{R}^3$ and rotation matrix $R \in \mathbb{R}^{3 \times 3}$ with $R S S^T R^T$.
For rendering, opacity $\sigma \in \mathbb{R}$ and color feature $F \in \mathbb{R}^C$ are associated to each 3D Gaussian, while $F$ is represented using spherical harmonics (SH) to model view-dependent color $c \in \mathbb{R}^3$. 
A tile-based rasterizer efficiently sorts the 3D Gaussians in front-to-back depth order and employs $\alpha$-blending, following projecting them onto the image plane as 2D Gaussians $G^{\prime}(x^{\prime})$ \cite{zwicker2001ewa}:
\begin{equation}
    C\left(x^{\prime}\right)=\sum_{i \in N}T_i c_i \sigma_i, \quad \sigma_i=\alpha_i G_i^{\prime}\left(x^{\prime}\right),
\end{equation}
where $x^{\prime}$ is the queried pixel, $N$ represents the number of sorted 2D Gaussians binded with that pixel, and $T$ denotes the transmittance as $\prod_{j=1}^{i-1}\left(1-\sigma_j\right)$.

\subsection{Scaffold-GS}
\label{sc-gs}
To efficiently manage Gaussian primitives, Scaffold-GS~\cite{lu2023scaffold} introduces anchors, each associated with a feature describing the local structure. From each anchor, $k$ neural Gaussians are emitted as follows:
\begin{equation}
\label{equ: offsets}
    \left\{\mu_0, \ldots, \mu_{k-1}\right\}
    = x_v+\left\{\mathcal{O}_0, \ldots, \mathcal{O}_{k-1}\right\} \cdot l_v
\end{equation}
where $x_v$ is the anchor position, $\{\mu_i\}$ denotes the positions of the i$^{th}$ neural Gaussian, and $l_v$ is a scaling factor controlling the predicted offsets $\{\mathcal{O}_i\}$.
In addition, opacities, scales, rotations, and colors are decoded from the anchor features through corresponding MLPs. For example, the opacities are computed as:
\begin{equation}
\label{equa: mlp}
    \{{\alpha}_{0}, ..., {\alpha}_{k-1}\} = \rm{F_{\alpha}}(\hat{f}_v, \Delta_{vc}, \vec{d}_{vc}),
\end{equation}
where $\{\alpha_i\}$ represents the opacity of the i$^{th}$ neural Gaussian, decoded by the opacity MLP $F_\alpha$. Here, $\hat{f}_v$, $\Delta_{vc}$, and $\vec{d}_{vc}$ correspond to the anchor feature, the relative viewing distance, and the direction to the camera, respectively. Once these properties are predicted, neural Gaussians are fed into the tile-based rasterizer, as described in~\cite{kerbl20233d}, to render images.
During the densification stage, Scaffold-GS treats anchors as the basic primitives. New anchors are established where the gradient of a neural Gaussian exceeds a certain threshold, while anchors with low average transparency are removed.
This structured representation improves robustness and storage efficiency compared to the vanilla 3D-GS.

\section{Methods}
\label{sec:method}

\begin{figure*}[t]
\includegraphics[width=\linewidth]{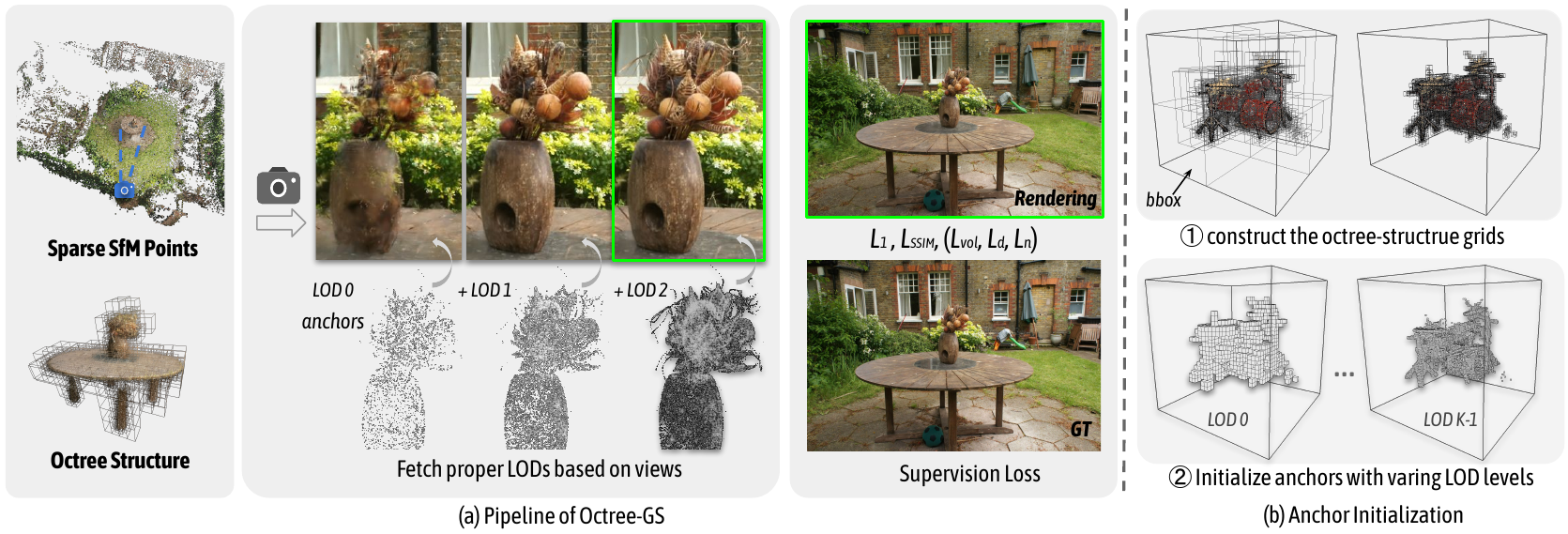}
\caption{
(a) Pipeline of \modelname: starting from given sparse SfM points, 
we construct octree-structured anchors from the bounded 3D space and assign them to the corresponding LOD level.
Unlike conventional 3D-GS methods treating all Gaussians equally, our approach involves primitives with varying LOD levels. 
We determine the required LOD levels based on the observation view 
and invoke corresponding anchors for rendering, as shown in the middle. 
As the LOD levels increase (from LOD $0$ to LOD $2$), the fine details of the vase accumulate progressively.
(b) Anchor Initialization: We construct the octree structure grids within the determined bounding box. Then, the anchors are initialized at the voxel center of each layer , with their LOD level corresponding to the octree layer of the voxel, ranging from $0$ to $K-1$.}
\label{fig:overview}
\vspace{-1em}
\end{figure*}

\modelname hierarchically organizes anchors into an octree structure to learn a neural scene from multiview images. Each anchor can emit different types of Gaussian primitives, such as explicit Gaussians~\cite{huang20242d, kerbl20233d} and neural Gaussians~\cite{lu2023scaffold}. By incorporating the octree structure, which naturally introduces a LOD hierarchy for both reconstruction and rendering, \modelname ensures consistently efficient training and rendering by dynamically selecting anchors from the appropriate LOD levels, allowing it to efficiently adapt to complex or large-scale scenes. Fig.~\ref{fig:overview} illustrates our framework. 

In this section, we first explain how to construct the octree from a set of given sparse SfM~\cite{schonberger2016structure} points in Sec.~\ref{method: lod_anchors}. Next, we introduce an adapted anchor densification strategy based on LOD-aware `growing' and `pruning' operations in Sec~\ref{method: anchor_refinement}. Sec.~\ref{method: progressive_training} then introduces a progressive training strategy that activates anchors from coarse to fine. Finally, to address reconstruction challenges in wild scenes, we introduce appearance embedding (Sec.~\ref{method: appearance_embedding}).

\subsection{LOD-structured Anchors}
\label{method: lod_anchors}

\subsubsection{Anchor Definition.} 
Inspired by Scaffold-GS~\cite{lu2023scaffold}, we introduce anchors to manage Gaussian primitives. These anchors are positioned at the centers of sparse, uniform voxel grids with varying voxel sizes. Specifically, anchors with higher LOD $L$ are placed within grids with smaller voxel sizes. 
In this paper, we define LOD 0 as the coarsest level. As the LOD level increases, more details are captured. Note that our LOD design is cumulative: the rendered images at LOD $K$ rasterize all Gaussian primitives from LOD $0$ to $K$.
Additionally, each anchor is assigned a LOD bias $\Delta L$ to account for local complexity, and each anchor is associated with $k$ Gaussian primitives for image rendering, whose positions are determined by Eq.~\ref{equ: offsets}. Moreover, our framework is generalized to support various types of Gaussians. For example, the Gaussian primitive can be explicitly defined with learnable distinct properties, such as 2D~\cite{huang20242d} or 3D Gaussians~\cite{kerbl20233d}, or they can be neural Gaussians decoded from the corresponding anchors, as described in Sec.~\ref{exp:framework}.

\subsubsection{Anchor Initialization.}
In this section, we describe the process of initializing octree-structured anchors from a set of sparse SfM points $\mathbf{P}$. First, the number of octree layers, $K$, is determined based on the range of observed distances. 
Specifically, we begin by calculating the distance $d_{ij}$ between each camera center of training image $i$ and SfM point $j$. The $r_d$th largest and $r_d$th smallest distances are then defined as $d_{max}$ and $d_{min}$, respectively. Here, $r_d$ is a hyperparameter used to discard outliers, which is typically set to $0.999$ in all our experiment. Finally, $K$ is calculated as:
\begin{equation}
    \begin{aligned}
        K &= \lfloor \log_{2}(\hat{d}_{max}/\hat{d}_{min}) \rceil + 1. \\
    \end{aligned}
\end{equation}
where $\lfloor \cdot \rceil$ denotes the round operator. The octree-structured grids with $K$ layers are then constructed, and the anchors of each layer are voxelized by the corresponding voxel size:
\begin{equation}
    \mathbf{V}_L =  \left\{ \left\lfloor\frac{\mathbf{P}}{\delta / 2^{L}}\right\rceil \cdot \delta / 2^{L} \right\},
\end{equation}
given the base voxel size $\delta$ for the coarsest layer corresponding to LOD 0 and $\mathbf{V}_L$ for initialed anchors in LOD $L$ . The properties of anchors and the corresponding Gaussian primitives are also initialized, please check the implementation~\ref{exp:framework} for details.

\subsubsection{Anchor Selection.} 
\label{sec: level_selection}

In this section, we explain how to select the appropriate visible anchors to maintain both stable real-time rendering speed and high rendering quality.
An ideal anchors is dynamically fetched from $K$ LOD levels based on the pixel footprint of projected Gaussians on the screen.
In practice, we simplify this by using the observation distance $d_{ij}$, as it is proportional to the footprint under consistent camera intrinsics. For varying intrinsics, a focal scale factor $s$ is applied to adjust the distance equivalently.
However, we find it sub-optimal if we estimate the LOD level solely based on observation distances. So we further set a learnable LOD bias $\Delta L$ for each anchor as a residual, which effectively supplements the high-frequency regions with more consistent details to be rendered during inference process, such as the presented sharp edges of an object as shown in Fig.~\ref{fig:ablate}.
In detail, for a given viewpoint $i$, the corresponding LOD level of an arbitrary anchor $j$ is estimated as:
\begin{equation}
    \hat{L_{ij}} = \lfloor L_{ij}^* \rfloor = \lfloor \Phi(\log_2(d_{max}/(d_{ij}*s))) + \Delta L_j\rfloor, 
\end{equation}
where $d_{ij}$ is the distance between viewpoint $i$ and anchor $j$. $\Phi(\cdot)$ is a clamping function that restricts the fractional LOD level $L_{ij}^*$ to the range $[0, K-1]$. Inspired by the progressive LOD techniques~\cite{hoppe2023progressive}, \modelname renders images using cumulative LOD levels rather than a single LOD level.
In summary, the anchor will be selected if its LOD level $L_j \leq \hat{L_{ij}}$. 
We iteratively evaluate all anchors and select those that meet this criterion, as illustrated in Fig.~\ref{fig:octree_init}. The Gaussian primitives emitted from the selected anchors are then passed into the rasterizer for rendering.

During inference, to ensure smooth rendering transitions between different LOD levels without introducing visible artifacts, we adopt an opacity blending technique inspired by~\cite{xu2023vr, xiangli2022bungeenerf}. We use piecewise linear interpolation between adjacent levels to make LOD transitions continuous, effectively eliminating LOD aliasing. Specifically, in addition to fully satisfied anchors, we also select nearly satisfied anchors that meet the criterion $L_j = \hat{L_{ij}} + 1$. The Gaussian primitives of these anchors are also passed to the rasterizer, with their opacities scaled by $L_{ij}^* - \hat{L_{ij}}$.

\subsection{Adaptive Anchor Gaussians Control}
\label{method: anchor_refinement}
\subsubsection{Anchor Growing.}

Following the approach of~\cite{kerbl20233d}, we use the view-space positional gradients of Gaussian primitives as a criterion to guide anchor densification. New anchors are grown in the unoccupied voxels across the octree-structured grids, following the practice of~\cite{lu2023scaffold}. Specifically, every $T$ iterations, we calculate the average accumulated gradient of the spawned Gaussian primitives, denoted as $\nabla_g$. Gaussian primitives with $\nabla_g$ exceeding a predefined threshold $\tau_g$ are considered significant and they are converted into new anchors if located in empty voxels. In the context of the octree structure, the question arises: which LOD level should be assigned to these newly converted anchors? To address this, we propose a `next-level' growing operation. This method adjusts the growing strategy by adding new anchors at varying granularities, with Gaussian primitives that have exceptionally high gradients being promoted to higher levels. To prevent overly aggressive growth into higher LOD levels, we monotonically increase the difficulty of growing new anchors to higher LOD levels by setting the threshold $\tau_g^{L}=\tau_g * 2^{\beta L}$, where $\tau_g$ and $\beta$ are both hyperparameters, with default values of $0.0002$ and $0.2$, respectively. Gaussians at level $L$ are only promoted to the next level $L+1$ if $\nabla_g > \tau_g^{L+1}$, and they remain at the same level if $\tau_g^{L} < \nabla_g < \tau_g^{L+1}$.

We also utilize the gradient as the complexity cue of the scene to adjust the LOD bias $\Delta L$. The gradient of an anchor is defined as the average gradient of the spawned Gaussian primitives, denoted as $\nabla_v$. We select those anchors with $\nabla_v > \tau_g^{L} * 0.25$, and increase the corresponding $\Delta L$ by a small user-defined quantity $\epsilon$: $\Delta L = \Delta L + \epsilon$. We empirically set $\epsilon = 0.01$.

\begin{figure*}[t!]
\includegraphics[width=\linewidth]{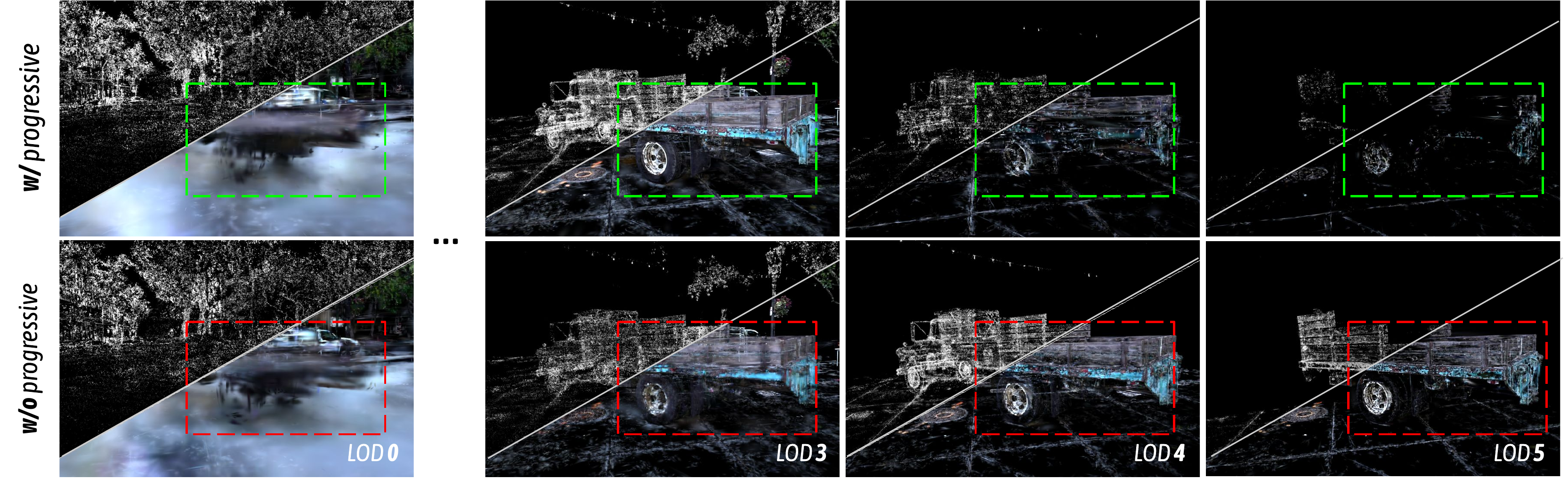}
\caption{
Visualization of anchors and projected 2D Gaussians in varying LOD levels.
(1) The first row depicts scene decomposition with our full model, employing a coarse-to-fine training strategy as detailed in Sec.~\ref{method: progressive_training}. A clear division of roles is evident between varying LOD levels: LOD 0 captures most rough contents, and higher LODs gradually recover the previously missed high-frequency details. This alignment with our motivation allows for more efficient allocation of model capacity with an adaptive learning process.
(2) In contrast, our ablated progressive training studies (elaborated in Sec.~\ref{sec:ablation}) take a naive approach. Here, all anchors are simultaneously trained, leading to an entangled distribution of Gaussian primitives across all LOD levels.
}
\label{fig:octree_init}
\vspace{-1em}
\end{figure*}

\subsubsection{Anchor Pruning.}
\label{method:anchor_prune}
To eliminate redundant and ineffective anchors, we compute the average opacity of Gaussians generated over $T$ training iterations, in a manner similar to the strategies adopted in~\cite{lu2023scaffold}.

\begin{figure}[t!]
\includegraphics[width=\linewidth]{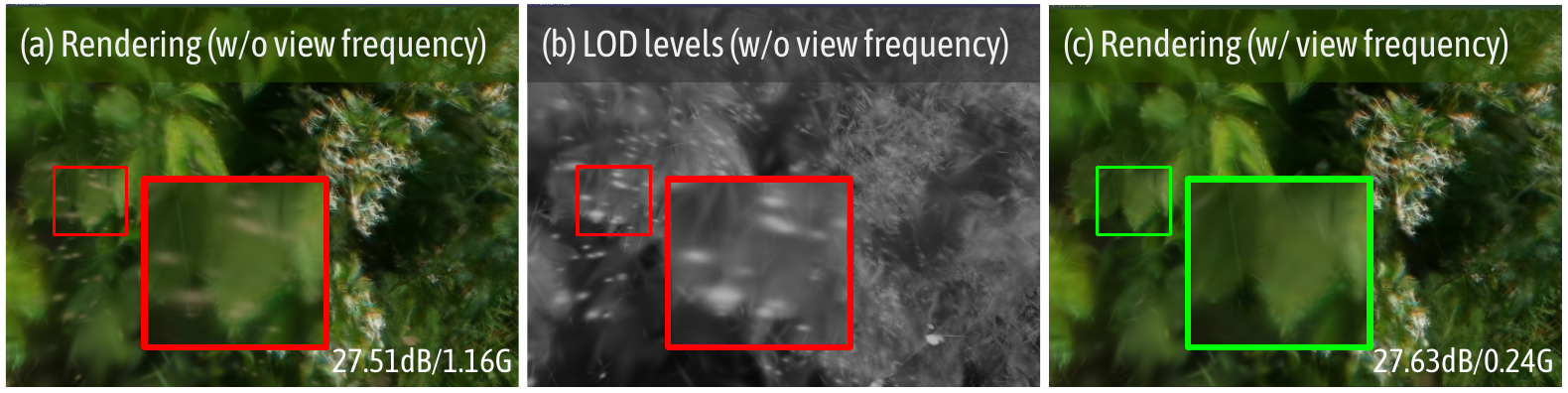}
\caption{Illustration of the effect of view frequency. We visualize the rendered image and the corresponding LOD levels (with whiter colors indicating higher LOD levels) from a novel view. We observe that insufficiently optimized anchors will produce artifacts if pruning is based solely on opacity. After pruning anchors based on view frequency, not only are the artifacts eliminated, but the final storage is also reduced. Last row metrics: PSNR/storage size.}
\centering
\label{fig:view_frequency}
\vspace{-1em}
\end{figure}

Moreover, we observe that some intolerable floaters appear in Fig.~\ref{fig:view_frequency} (a) because a significant portion of anchors are not visible or selected in most training view frustums. Consequently, they are not sufficiently optimized, impacting rendering quality and storage overhead significantly. To address this issue, we define `view-frequency' as the probability that anchors are selected in the training views, which directly correlates with the received gradient. We remove anchors with the view-frequency below $\tau_v$, where $\tau_v$ represents the visibility threshold. This strategy effectively eliminates floaters, improving visual quality and significantly reducing storage, as demonstrated in Fig.~\ref{fig:view_frequency}.

\subsection{Progressive Training}
\label{method: progressive_training}

Optimizing anchors across all LOD levels simultaneously poses inherent challenges in explaining rendering with decomposed LOD levels. All LOD levels try their best to represent the 3D scene, making it difficult to decompose them thus leading to large overlaps.

Inspired by the progressive training strategy commonly used in prior NeRF methods~\cite{park2021nerfies,xiangli2022bungeenerf,li2023neuralangelo}, we implement a coarse-to-fine optimization strategy. 
begins by training on a subset of anchors representing lower LOD levels and progressively activates finer LOD levels throughout optimization, complementing the coarse levels with fine-grained details. 
In practice, we iteratively activate an additional LOD level after $N$ iterations. Empirically, we start training from $\lfloor \frac{K}{2} \rfloor$ level to balance visual quality and rendering efficiency. Additionally, more time is dedicated to learning the overall structure because we want coarse-grained anchors to perform well in reconstructing the scene as the viewpoint moves away. Therefore, we set $N_{i-1} = \omega N_{i}$, where $N_i$ denotes the training iterations for LOD level $L=i$, and $\omega \geq 1$ is the growth factor. 
Note that during the progressive training stage, we disable the next level grow operator.

With this approach, we find that the anchors can be arranged more faithfully into different LOD levels as demonstrated in Fig.~\ref{fig:octree_init}, reducing anchor redundance and leading to faster rendering without reducing the rendering quality.

\subsection{Appearance Embedding}
\label{method: appearance_embedding}
In large-scale scenes, the exposure compensation of training images is always inconsistent, and 3D-GS~\cite{kerbl20233d} tends to produce artifacts by averaging the appearance variations across training images. To address this, and following the approach of prior NeRF papers~\cite{martin2021nerf, tancik2022block}, we integrate Generative Latent Optimization (GLO)~\cite{bojanowski2017optimizing} to generate the color of Gaussian primitives. For instance, we introduce a learnable individual appearance code for each anchor, which is fed as an addition input to the color MLP to decode the colors of the Gaussian primitives. This allows us to effectively model in-the-wild scenes with varying appearances. Moreover, we can also interpolate the appearance code to alter the visual appearance of these environments, as shown in Fig.~\ref{fig:ape}.

\section{Experiments}
\label{sec:exp}

\begin{table*}[htbp]
\centering
\renewcommand{\arraystretch}{1.15}
\setlength{\tabcolsep}{2pt}
\caption{Quantitative comparison on real-world datasets~\cite{barron2022mip,knapitsch2017tanks,hedman2018deep}. \modelname consistently achieves superior rendering quality compared to baselines with reduced number of Gaussian primitives rendered per-view. We highlight \textbf{best} and \underline{second-best} in each category.}
\label{tab:real_q}
\resizebox{1\linewidth}{!}{
\begin{tabular}{l|cccc|cccc|cccc}
\toprule
Dataset & \multicolumn{4}{c|}{Mip-NeRF360} & \multicolumn{4}{c|}{Tanks\&Temples} & \multicolumn{4}{c}{Deep Blending} \\
\begin{tabular}{c|c} Method & Metrics \end{tabular}  & PSNR\(\uparrow\) & SSIM\(\uparrow\) & LPIPS\(\downarrow\) & \#GS(k)/Mem & PSNR\(\uparrow\) & SSIM\(\uparrow\) & LPIPS\(\downarrow\) & \#GS(k)/Mem & PSNR\(\uparrow\) & SSIM\(\uparrow\) & LPIPS\(\downarrow\) & \#GS(k)/Mem \\
\midrule

Mip-NeRF360~\cite{barron2022mip} & 27.69 & 0.792 & 0.237 & - & 23.14 & 0.841 & 0.183 & - & 29.40 & 0.901 & 0.245 & - \\

2D-GS~\cite{kerbl20233d} & 26.93 & 0.800 & 0.251 & \underline{397}/440.8M & 23.25 & 0.830 & 0.212 & \underline{352}/204.4M & 29.32 & 0.899 & 0.257 & 196/335.3M \\

3D-GS~\cite{kerbl20233d} & 27.54 & 0.815 & 0.216 & 937/786.7M & 23.91 & 0.852 & 0.172 & 765/430.1M & 29.46 & 0.903 & 0.242 & 398/705.6M \\

Mip-Splatting~\cite{yu2023mip} & 27.61 & \underline{0.816} & \underline{0.215} & 1013/838.4M & 23.96 & 0.856 & 0.171 & 832/500.4M & 29.56 & 0.901 & 0.243 & 410/736.8M \\

Scaffold-GS~\cite{lu2023scaffold} & \underline{27.90} & 0.815 & 0.220 & 666/\underline{197.5M} & \underline{24.48} & \underline{0.864} & \underline{0.156} & 626/\underline{167.5M} & \underline{30.28} & \underline{0.909} & \textbf{0.239} & 207/\underline{125.5M} \\

\hline

Anchor-2D-GS & 26.98 & 0.801 & 0.241 & 547/392.7M & 23.52 & 0.835 & 0.199 & 465/279.0M & 29.35 & 0.896 & 0.264 & 162/289.0M \\

Anchor-3D-GS & 27.59 & 0.815 & 0.220 & 707/492.0M & 24.02 & 0.847 & 0.184 & 572/349.2M & 29.66 & 0.899 & 0.260 & 150/272.9M\\

\hline

Our-2D-GS & 27.02 & 0.801 & 0.241 & \textbf{397}/371.6M & 23.62 & 0.842 & 0.187 & \textbf{330}/191.2M & 29.44 & 0.897 & 0.264 & \underline{84}/202.3M \\

Our-3D-GS & 27.65 & 0.815 & 0.220 & 504/418.6M & 24.17 & 0.858 & 0.161 & 424/383.9M & 29.65 & 0.901 & 0.257 & \textbf{79}/180.0M \\

Our-Scaffold-GS & \textbf{28.05} & \textbf{0.819} & \textbf{0.214} & 657/\textbf{139.6M} & \textbf{24.68} & \textbf{0.866} & \textbf{0.153} & 443/\textbf{88.5M} & \textbf{30.49} & \textbf{0.912} & \underline{0.241} & 112/\textbf{71.7M} \\

\bottomrule
\end{tabular}}
\vspace{-1em}
\end{table*}

\begin{figure*}[t!]
\centering
\includegraphics[width=\linewidth]{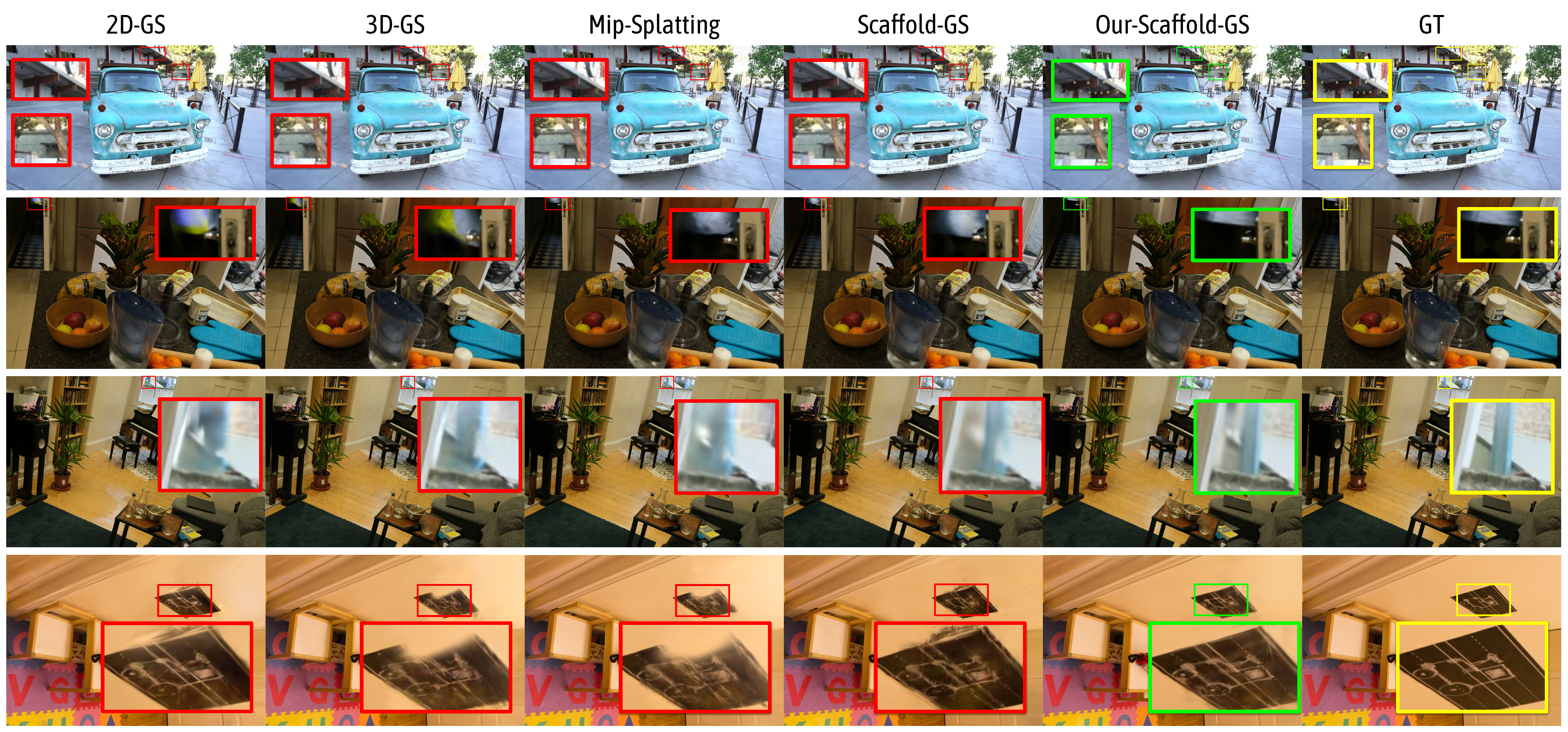}
\vspace{-10pt}
\caption{Qualitative comparison of our method and SOTA methods~\cite{huang20242d,kerbl20233d,yu2023mip,lu2023scaffold} across diverse datasets~\cite{barron2022mip,knapitsch2017tanks,hedman2018deep, xiangli2022bungeenerf}. We highlight the difference with colored patches.
Compared to existing baselines, our method successfully captures very fine details presented in indoor and outdoor scenes, particularly for objects with thin structures such as trees, light-bulbs, decorative texts and etc..
}
\label{fig:Rendering_comparison}
\centering
\vspace{-1.5em}
\end{figure*}

\begin{table*}[htbp]
\renewcommand{\arraystretch}{1.15}
\setlength{\tabcolsep}{3pt}
\centering
\caption{Quantitative comparison on large-scale urban dataset~\cite{li2023matrixcity,turki2022mega,lin2022capturing}. In addition to three methods compared in Tab.~\ref{tab:real_q}, we also compare our method with CityGaussian~\cite{liu2024citygaussian} and Hierarchical-GS~\cite{kerbl2024hierarchical}, both of which are specifically targeted at large-scale scenes. It is evident that \modelname outperforms the others in both rendering quality and storage efficiency.
We highlight \textbf{best} and \underline{second-best} in each category.}
\resizebox{1\linewidth}{!}{
\begin{tabular}{l|cccc|cccc|cccc}
\toprule
Dataset & \multicolumn{4}{c|}{Block\_Small} & \multicolumn{4}{c|}{Block\_All} & \multicolumn{4}{c}{Building} \\ 
\begin{tabular}{c|c} Method & Metrics \end{tabular}  & PSNR\(\uparrow\) & SSIM\(\uparrow\) & LPIPS\(\downarrow\) & \#GS(k)/Mem & PSNR\(\uparrow\) & SSIM\(\uparrow\) & LPIPS\(\downarrow\) & \#GS(k)/Mem & PSNR\(\uparrow\) & SSIM\(\uparrow\) & LPIPS\(\downarrow\) & \#GS(k)/Mem \\
\midrule

3D-GS~\cite{kerbl20233d} & 26.82 & 0.823 & 0.246 & 1432/3387.4M & 24.45 & 0.746 & 0.385 & 979/3584.3M & 22.04 & 0.728 & 0.332 & 842/1919.2M \\ 

Mip-Splatting~\cite{yu2023mip} & 27.14 & 0.829 & 0.24 & 860/3654.6M & 24.28 & 0.742 & 0.388 & 694/3061.8M & 22.13 & 0.726 & 0.335 & 1066/2498.6M \\ 

Scaffold-GS~\cite{lu2023scaffold} & 29.00 & 0.868 & 0.210 & 357/\textbf{371.2M} & 26.30 & 0.808 & 0.293 & 690/\underline{2272.2M} & 22.42 & 0.719 & 0.336 & \textbf{438}/\textbf{833.2M} \\ 

\hline

CityGaussian~\cite{liu2024citygaussian} & 27.46 & 0.808 & 0.267 & 538/4382.7M & 26.26 & 0.800 & 0.324 & 235/4316.6M & 20.94 & 0.706 & 0.310 & 520/3026.8M\\ 

Hierarchical-GS~\cite{kerbl2024hierarchical} & 27.69 & 0.823 & 0.276 & 271/1866.7M & 26.00 & 0.803 & 0.306 & 492/4874.2M & \underline{23.28} & \underline{0.769} & \underline{0.273} & 1973/3778.6M \\ 

Hierarchical-GS($\tau_1$) & 27.67 & 0.823 & 0.276 & 271/1866.7M & 25.44 & 0.788 & 0.320 & 435/4874.2M& 23.08 & 0.758 & 0.285 & 1819/3778.6M \\ 

Hierarchical-GS($\tau_2$) & 27.54 & 0.820 & 0.280 & 268/1866.7M & 25.39 & 0.783 & 0.325 & 355/4874.2M& 22.55 & 0.726 & 0.313 & 1473/3778.6M \\ 

Hierarchical-GS($\tau_3$) & 26.60 & 0.794 & 0.319 & \underline{221}/1866.7M & 25.19 & 0.773 & 0.352 & \textbf{186}/4874.2M & 21.35 & 0.635 & 0.392 & 820/3778.6M\\ 
\hline

Our-3D-GS & \underline{29.37} & \underline{0.875} & \underline{0.197} & \textbf{175}/755.7M & \underline{26.86} & \underline{0.833} & \underline{0.260} & \underline{218}/3205.1M & 22.67 & 0.736 & 0.320 & \underline{447}/1474.5M 
\\

Our-Scaffold-GS & \textbf{29.83} & \textbf{0.887} & \textbf{0.192} & 360/\underline{380.3M} & \textbf{27.31} & \textbf{0.849} & \textbf{0.229} & 344/\textbf{1648.6M} & \textbf{23.66} & \textbf{0.776} & \textbf{0.267} & 619/\underline{1146.9M} \\

\bottomrule
\end{tabular}
}

\vspace{0.5em}
\resizebox{\linewidth}{!}{
\begin{tabular}{l|cccc|cccc|cccc}
\toprule
Dataset & \multicolumn{4}{c|}{Rubble} & \multicolumn{4}{c|}{Residence} & \multicolumn{4}{c}{Sci-Art} \\
\begin{tabular}{c|c} Method & Metrics \end{tabular}  & PSNR\(\uparrow\) & SSIM\(\uparrow\) & LPIPS\(\downarrow\) & \#GS(k)/Mem & PSNR\(\uparrow\) & SSIM\(\uparrow\) & LPIPS\(\downarrow\) & \#GS(k)/Mem & PSNR\(\uparrow\) & SSIM\(\uparrow\) & LPIPS\(\downarrow\) & \#GS(k)/Mem \\
\midrule

3D-GS~\cite{kerbl20233d} & 25.20 & 0.757 & 0.318 & 956/2355.2M & 21.94 & \underline{0.764} & 0.279 & 1209/2498.6M & 21.85 & 0.787 & 0.311 & 705/950.6M \\ 

Mip-Splatting~\cite{yu2023mip} & 25.16 & 0.746 & 0.335 & 760/1787.0M & 21.97 & 0.763 & 0.283 & 1301/2570.2M & 21.92 & 0.784 & 0.321 & 615/880.2M \\ 

Scaffold-GS~\cite{lu2023scaffold}  & 24.83 & 0.721 & 0.353 & \underline{492}/\textbf{470.3M} & \underline{22.00} & 0.761 & 0.286 & 596/\underline{697.7M} & 22.56 & 0.796 & 0.302 & \underline{526}/\textbf{452.5M} \\ 

\hline

CityGaussian~\cite{liu2024citygaussian} & 24.67 & 0.758 & \textbf{0.286} & 619/3000.3M & 21.92 & \textbf{0.774} & \textbf{0.257} & 732/3196.0M & 20.07 & 0.757 & 0.290 & \textbf{461}/1300.3M \\ 

Hierarchical-GS~\cite{kerbl2024hierarchical} & \textbf{25.37} & \underline{0.761} & 0.300 & 1541/2345.0M & 21.74 & 0.758 & \underline{0.274} & 2040/2498.6M & 22.02 & 0.810 & 0.257 & 2363/2160.6M \\ 

Hierarchical-GS($\tau_1$) & 25.27 & 0.754 & 0.305 & 1478/2345.0M & 21.70 & 0.756 & 0.276 & 1972/2498.6M & 22.00 & 0.808 & 0.259 & 2226/2160.6M \\ 

Hierarchical-GS($\tau_2$) & 24.80 & 0.724 & 0.329 & 1273/2345.0M & 21.49 & 0.743 & 0.291 & 1694/2498.6M & 21.93 & 0.802 & 0.268 & 1916/2160.6M \\ 

Hierarchical-GS($\tau_3$) & 23.55 & 0.628 & 0.414 & 781/2345.0M & 20.69 & 0.683 & 0.363 & 976/2498.6M & 21.50 & 0.766 & 0.324 & 1165/2160.6M \\ 

\hline

Our-3D-GS & 24.67 & 0.728 & 0.345 & \textbf{489}/1392.6M & 21.60 & 0.736 & 0.314 & \underline{350}/986.2M & 22.52 & 0.817 & 0.256 & 630/1331.2M \\

Our-Scaffold-GS & \underline{25.34} & \textbf{0.763} & \underline{0.299} & 674/\underline{693.5M} & \textbf{22.29} & 0.762 & 0.288 & \textbf{344}/\textbf{618.8M} & \textbf{23.38} & \textbf{0.828} & \textbf{0.240} & 871/\underline{866.9M} \\

\bottomrule

\end{tabular}
}

\vspace{-1em}
\label{tab:large_scale}
\end{table*}

\begin{figure*}[t!]
\centering
\includegraphics[width=\linewidth]{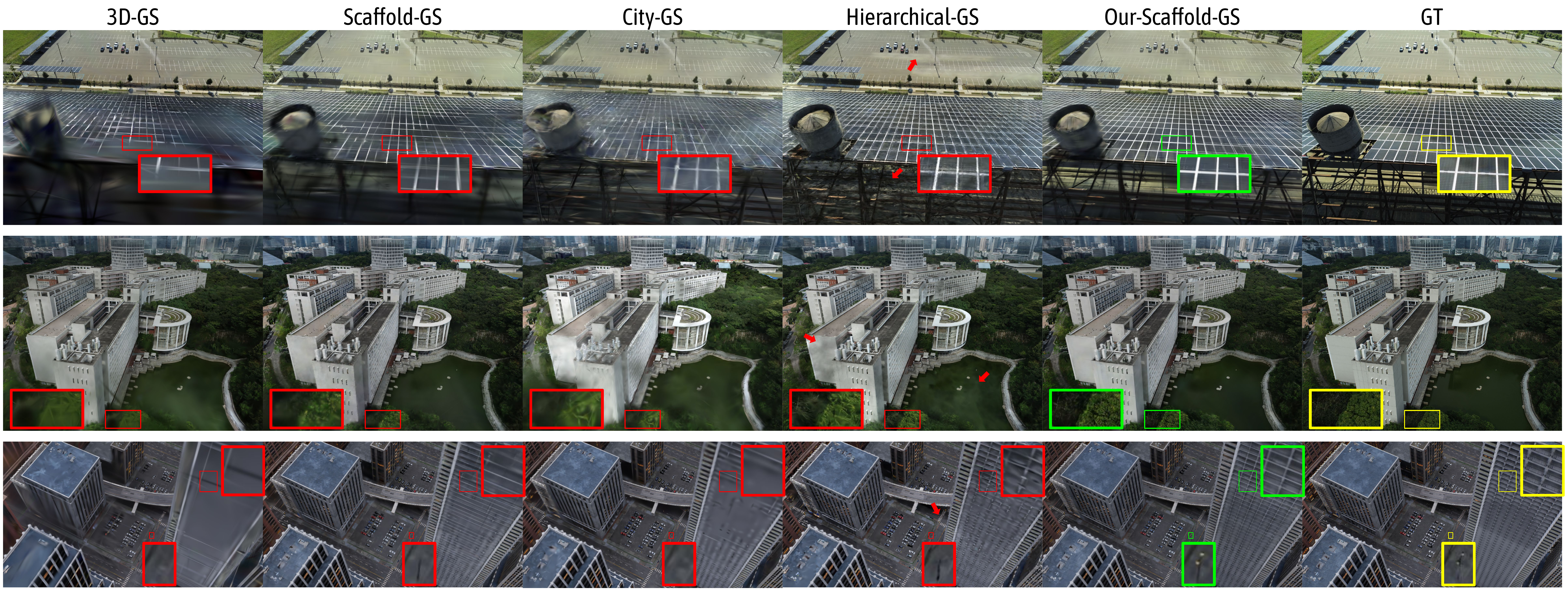}
\caption{Qualitative comparisons of \modelname against baselines~\cite{kerbl20233d,lu2023scaffold,liu2024citygaussian,kerbl2024hierarchical} across large-scale datasets~\cite{turki2022mega,lin2022capturing,li2023matrixcity}. 
As shown in the highlighted patches and arrows above, our method consistently outperforms the baselines, especially in modeling fine details (1st \& 3rd row), texture-less regions (2nd row), which are common in large-scale scenes.}
\vspace{-1em}
\label{fig:large_scale}
\end{figure*}

\subsection{Experimental Setup}
\subsubsection{Datasets}

We conduct comprehensive evaluations on $21$ small-scale scenes and $7$ large-scale scenes from various 
public datasets. Small-scale scenes include 9 scenes from Mip-NeRF360~\cite{barron2022mip}, 2 scenes from Tanks$\&$Temples~\cite{knapitsch2017tanks}, 2 scenes in DeepBlending~\cite{hedman2018deep} and 8 scenes from BungeeNeRF~\cite{xiangli2022bungeenerf}.

For large-scale scenes, we provide a detailed explanation.
Specifically, we evaluate on the Block\_Small and Block\_All scenes (the latter being 10$\times$ larger) in the MatrixCity~\cite{li2023matrixcity} dataset, which uses Zig-Zag trajectories commonly used in oblique photography.
%
In the MegaNeRF~\cite{turki2022mega} dataset, we choose the Rubble and Building scenes, while in the UrbanScene3D~\cite{lin2022capturing} dataset, we select the Residence and Sci-Art scenes. Each scene contains thousands of high-resolution images, and we use COLMAP~\cite{schonberger2016structure} to obtain sparse SfM points and camera poses.
In the Hierarchical-GS~\cite{kerbl2024hierarchical} dataset, we maintain their original settings and compare both methods on a chunk of the SmallCity scene, which includes 1,470 training images and 30 test images, each paired with depth and mask images.

For the Block\_All scene and the SmallCity scene, we employ the train and test information provided by their authors. For other scenes, we uniformly select one out of every eight images as test images, with the remaining images used for training. 
%


\subsubsection{Metrics}

In addition to the visual quality metrics PSNR, SSIM~\cite{wang2004image} and LPIPS~\cite{zhang2018unreasonable}, we also report the file size for storing anchors, the average selected Gaussian primitives used in per-view rendering process, and the rendering speed FPS as a fair indicator for memory and rendering efficiency. 
We provide the average quantitative metrics on test sets in the main paper and leave the full table for each scene in the supplementary material.

\subsubsection{Baselines}
We compare our method against 2D-GS~\cite{huang20242d}, 3D-GS~\cite{kerbl20233d}, Scaffold-GS~\cite{lu2023scaffold}, Mip-Splatting~\cite{yu2023mip} and two concurrent works, CityGaussian~\cite{liu2024citygaussian} and Hierarchical-GS~\cite{kerbl2024hierarchical}. 
In the Mip-NeRF360\cite{barron2022mip}, Tanks$\&$Temples\cite{knapitsch2017tanks}, and DeepBlending\cite{hedman2018deep} datasets, we compare our method with the top four methods. In the large-scale scene datasets MatrixCity~\cite{li2023matrixcity}, MegaNeRF~\cite{turki2022mega} and UrbanScene3D~\cite{lin2022capturing}, we add the results of CityGaussian and Hierarchical-GS for comparison. To ensure consistency, we remove depth supervision from Hierarchical-GS in these experiments. Following the original setup of Hierarchical-GS, we report results at different granularities (leaves, $\tau_1=3$, $\tau_2=6$, $\tau_3=15$), each one is after the optimization of the hierarchy.
In the street-view dataset, we compare exclusively with Hierarchical-GS, the current state-of-the-art (SOTA) method for street-view data. In this experiment, we apply the same depth supervision used in Hierarchical-GS for fair comparison.

\subsubsection{Instances of Our Framework}
\label{exp:framework}

To demonstrate the generalizability of the proposed framework, we apply it to 2D-GS~\cite{huang20242d}, 3D-GS~\cite{kerbl20233d}, and Scaffold-GS~\cite{lu2023scaffold}, which we refer to as Our-2D-GS, Our-3D-GS and Our-Scaffold-GS, respectively. In addition, for a fair comparison and deeper analysis, we modify 2D-GS and 3D-GS to anchor versions. Specifically, we voxelize the input SfM points to anchors and assign each of them 2D or 3D Gaussians, while maintaining the same densification strategy as Scaffold-GS. We denote these modified versions as Anchor-2D-GS and Anchor-3D-GS.

\subsubsection{Implementation Details}
For 3D-GS model we employ standard L1 and SSIM loss, with weights set to 0.8 and 0.2, respectively.
For 2D-GS model, we retain the distortion loss $\mathcal{L}_d=\sum_{i, j} \omega_i \omega_j\left|z_i-z_j\right|$ and normal loss $\mathcal{L}_n=\sum_i \omega_i\left(1-\mathbf{n}_i^{\mathrm{T}} \mathbf{N}\right)$ , with weights set to 0.01 and 0.05, respectively.
For Scaffold-GS model, we keep an additional volume regularization loss $\mathcal{L}_{\mathrm{vol}}=\sum_{i=1}^{N} \operatorname{Prod}\left(s_i\right)$, with a weight set to 0.01.

We adjust the training and densification iterations across all compared methods to ensure a fair comparison. Specifically, for small-scale scenes~\cite{barron2022mip, knapitsch2017tanks, hedman2018deep, xiangli2022bungeenerf, kerbl2024hierarchical}, training was set to 40k iterations, with densification concluding at 20k iterations. For large-scale scenes~\cite{li2023matrixcity, turki2022mega, lin2022capturing}, training was set to 100k iterations, with densification ending at 50k iterations.

We set the voxel size to $0.001$ for all scenes in the modified anchor versions of 2D-GS~\cite{huang20242d}, 3D-GS~\cite{kerbl20233d}, and Scaffold-GS~\cite{lu2023scaffold}, while for our method, we set the voxel size for the intermediate level of the anchor grid to $0.02$. For the progress training, we set the total training iteration to $10$k with $\omega = 1.5$. Since not all layers are fully densified during the progressive training process, we extend the densification by an additional $10$k iterations, and we set the densification interval $T=100$ empirically. We set the visibility threshold $\tau_v$ to $0.7$ for the small-scale scenes~\cite{barron2022mip, knapitsch2017tanks, hedman2018deep, xiangli2022bungeenerf},as these datasets contain densely captured images, while for large-scale scenes~\cite{turki2022mega, lin2022capturing, kerbl2024hierarchical}, we set $\tau_v$ to $0.01$. In addition, for the multi-scale dataset~\cite{xiangli2022bungeenerf}, we set $\tau_v$ to $0.2$. 

All experiments are conducted on a single NVIDIA A100 80G GPU. To avoid the impact of image storage on GPU memory, all images were stored on the CPU.

\begin{figure*}[htbp]
\centering
\includegraphics[width=\linewidth]{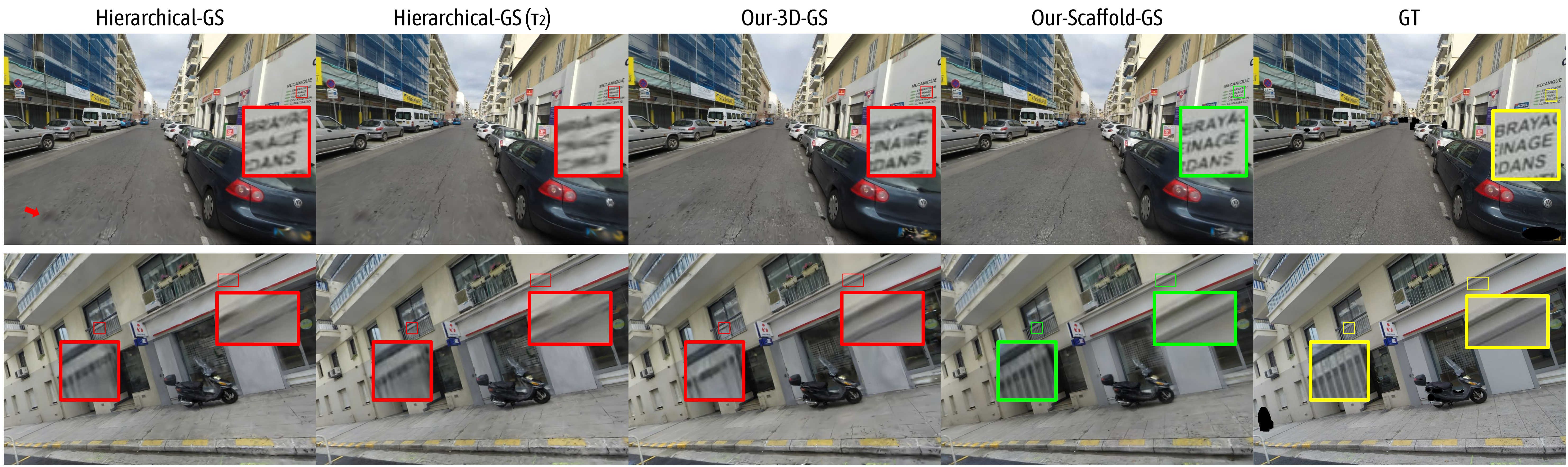}
\caption{Qualitative comparisons of our approach against Hierarchical-GS~\cite{kerbl2024hierarchical}. We present both the highest-quality setting (leaves) and a reasonably reduced LOD setting ($\tau_2$ = 6 pixels). \modelname demonstrates superior performance in street views, specially in thin geometries and texture-less regions (\eg, railings, signs and pavements.)}
\label{fig:street_view}
\vspace{-1em}
\end{figure*}

\begin{table}[t!]
\centering
\renewcommand{\arraystretch}{1.15}
\setlength{\tabcolsep}{3pt}
\caption{Quantitative comparison on the SMALLCITY scene of the Hierarchical-GS~\cite{kerbl2024hierarchical} dataset. The competing metrics are sourced from the original paper.
}
\vspace{-3pt}
\label{tab:street_view}
\resizebox{\linewidth}{!}{
\begin{tabular}{l|cccc}
\toprule
Method & PSNR($\uparrow$) & SSIM($\uparrow$) & LPIPS($\downarrow$) & FPS($\uparrow$)\\

\midrule
3D-GS~\cite{kerbl20233d} & 25.34 & 0.776 & 0.337 & 99 \\

Hierarchical-GS~\cite{kerbl2024hierarchical} & \textbf{26.62} & 0.820 & 0.259 & 58 \\

Hierarchical-GS($\tau_1$) & \underline{26.53} & \underline{0.817} & \underline{0.263} &  86 \\

Hierarchical-GS($\tau_2$) & 26.29 & 0.810 & 0.275 & 110 \\

Hierarchical-GS($\tau_3$) & 25.68 & 0.786 & 0.324  & \textbf{159} \\

\midrule

Our-3D-GS & 25.77 & 0.811 & 0.272 & \underline{130} \\

Our-Scaffold-GS & 26.10 & \textbf{0.826} & \textbf{0.235} & 89 \\

\bottomrule
\end{tabular}}
\vspace{-2em}
\end{table}

\begin{figure}[t!]
\centering
\includegraphics[width=\linewidth]{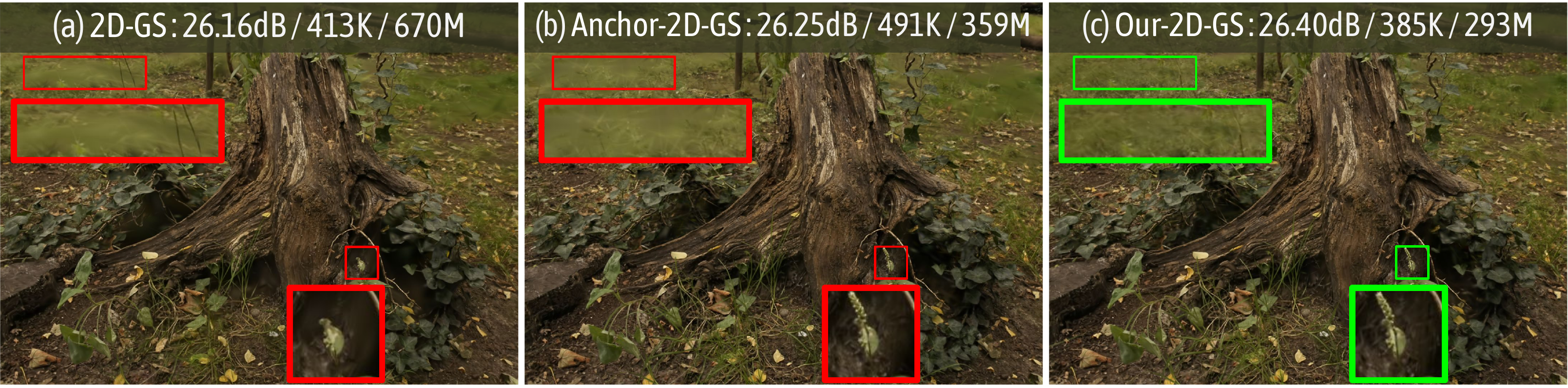}
\caption{Comparison of different versions of the 2D-GS~\cite{huang20242d} model. We showcase the rendering results on the stump scene from the Mip-NeRF360~\cite{barron2022mip} dataset. We report PSNR, average number of Gaussians for rendering and storage size.
}
\vspace{-1em}
\label{fig:anchor_LOD}
\end{figure}

\begin{table*}[htbp]
\centering
\renewcommand{\arraystretch}{1.15}
\setlength{\tabcolsep}{3pt}
\caption{
Quantitative comparison on the BungeeNeRF~\cite{xiangli2022bungeenerf} dataset. 
We provide metrics for each scale and their average across all four. Scale-1 denotes the closest views, while scale-4 covers the entire landscape. We note a notable rise in Gaussian counts for baseline methods when zooming out from scale 1 to 4, whereas our method maintains a significantly lower count, ensuring consistent rendering speed across all LOD levels. We highlight \textbf{best} and \underline{second-best} in each category.}
\label{tab:bungeenerf}
\resizebox{1\linewidth}{!}{
\begin{tabular}{l|cccc|cc|cc|cc|cc}
\toprule
Dataset & \multicolumn{4}{c|}{BungeeNeRF (Average)} & \multicolumn{2}{c|}{scale-1} & \multicolumn{2}{c|}{scale-2} & \multicolumn{2}{c|}{scale-3} & \multicolumn{2}{c}{scale-4} \\

\begin{tabular}{c|c} Method & Metrics \end{tabular}  & PSNR\(\uparrow\) & SSIM\(\uparrow\) & LPIPS\(\downarrow\) & \#GS(k)/Mem & PSNR\(\uparrow\)  & \#GS(k) & PSNR\(\uparrow\) & \#GS(k) & PSNR\(\uparrow\) & \#GS(k) & PSNR\(\uparrow\) & \#GS(k) \\
\midrule

2D-GS~\cite{kerbl20233d} & 27.10 & 0.903 & 0.121 & 1079/886.1M & 28.18 & \textbf{205} & 28.11 & \textbf{494} & 25.99 & 1826 & 23.71 & 2365 \\

3D-GS~\cite{kerbl20233d} & 27.79 & 0.917 & \underline{0.093} & 2686/1792.3M & 30.00 & 522 & 28.97 & 1272 & 26.19 & 4407 & 24.20 & 5821 \\

Mip-Splatting~\cite{yu2023mip} & 28.14 & \underline{0.918} & 0.094 & 2502/1610.2M & 29.79 & 503 & 29.37 & 1231 & \textbf{26.74} & 4075 & 24.44 & 5298 \\

Scaffold-GS~\cite{lu2023scaffold} & \underline{28.16} & 0.917 & 0.095 & 1652/\underline{319.2M} & 30.48 & 303 & 29.18 & 768 & \underline{26.56} & 2708 & \underline{24.95} & 3876 \\

\hline

Anchor-2D-GS & 27.18 & 0.885 & 0.140 & 1050/533.8M & 29.80 & 260 & 28.26 & 601 & 25.43 & 1645 & 23.71 & 2026 \\

Anchor-3D-GS & 27.90 & 0.909 & 0.114 & 1565/790.3M & 30.85 & 391 & 29.29 & 905 & 26.13 & 2443 & 24.49 & 3009 \\

\hline

Our-2D-GS & 27.34 & 0.893 & 0.129 & \textbf{676}/736.1M & 30.09 & \underline{249} & 28.72 & \underline{511} & 25.42 & \textbf{1003} & 23.41 & \textbf{775} \\

Our-3D-GS & 27.94 & 0.909 & 0.110 & \underline{952}/1045.7M & \underline{31.11} & 411 & \underline{29.42} & 819 & 25.88 & \underline{1275} & 23.77 & \underline{938} \\

Our-Scaffold-GS & \textbf{28.39} & \textbf{0.923} & \textbf{0.088} & 1474/\textbf{296.7M} & \textbf{31.11} & 486 & \textbf{29.59} & 1010 & 26.51 & 2206 & \textbf{25.07} & 2167 \\

\bottomrule
\end{tabular}}
\end{table*}

\subsection{Results Analysis}
Our evaluation encompasses a wide range of scenes, including indoor and outdoor environments, both synthetic and real-world, as well as large-scale  urban scenes from both aerial views and street views. We demonstrate that our method preserves fine-scale details while reducing the number of Gaussians, resulting in faster rendering speed and lower storage overhead, as shown in Fig.~\ref{fig:Rendering_comparison},~\ref{fig:large_scale},~\ref{fig:street_view},~\ref{fig:anchor_LOD} and Tab.~\ref{tab:real_q},~\ref{tab:bungeenerf},~\ref{tab:large_scale},~\ref{tab:street_view},~\ref{tab:matrix_traj}.

\subsubsection{Performance Analysis}
\paragraph{Quality Comparisons}
Our method introduces anchors with octree structure, which decouple multi-scale Gaussian primitives into varying LOD levels. This approach enables finer Gaussian primitives to capture scene details more accurately, thereby enhancing the overall rendering quality. In Fig.~\ref{fig:Rendering_comparison},~\ref{fig:large_scale},~\ref{fig:street_view} and Tab.~\ref{tab:real_q},~\ref{tab:large_scale},~\ref{tab:street_view}, we compare \modelname to previous state-of-the-art (SOTA) methods, demonstrating that our method consistently outperforms the baselines across both small-scale and large-scale scenes, especially in fine details and texture-less regions. Notably, when compared to Hierarchical-GS~\cite{kerbl2024hierarchical} on the street-view dataset, \modelname exhibits slightly lower PSNR values but significantly better visual quality, with LPIPS scores of 0.235 for ours and 0.259 for theirs. 

\paragraph{Storage Comparisons}
As shown in Tab.~\ref{tab:real_q},~\ref{tab:large_scale}, our method reduces the number of Gaussian primitives used for rendering, resulting in faster rendering speed and lower storage overhead. This demonstrates the benefits of our two main improvements: 1) our LOD structure efficiently arranges Gaussian primitives, with coarse primitives representing low-frequency scene information, which previously required redundant primitives; and 2) our view-frequency strategy significantly prunes unnecessary primitives.

\paragraph{Variants Comparisons}
As described in Sec.~\ref{sec:method}, our method is agnostic to the specific Gaussian representation and can be easily adapted to any Gaussian-based method with minimal effort. In Tab.~\ref{tab:real_q}, the modified anchor-version of 2D-GS\cite{huang20242d} and 3D-GS~\cite{kerbl20233d} achieve competitive rendering quality with fewer file storage than the original methods. This demonstrates that the anchor design organizes the Gaussian primitives more efficiently, reducing redundancy and creating a more compact way. More than the anchor design, \modelname delivers better visual performance and fewer Gaussian primitives as shown in Tab.~\ref{tab:real_q}, which benefits from the explicit, multi-level anchor design. 
In Fig.~\ref{fig:anchor_LOD}, we compare the vanilla 2D-GS with the anchor-version and octree-version method. Among them, the octree-version provides the most detail and the least amount of Gaussian primitives and storage.

\begin{table}[t!]
\centering
\caption{Quantitative comparison of rendering speed on the MatrixCity~\cite{li2023matrixcity} dataset. We report the averaged FPS on three novel view trajectories (Fig. \ref{fig:traj}). Our method shows consistent rendering speed above $30$ FPS at $2k$ image resolution while all baseline methods fail to meet the real-time performance.}
\label{tab:matrix_traj}
\resizebox{0.8\linewidth}{!}{
\begin{tabular}{l|ccc}
\toprule
\begin{tabular}{c|c} Method & Traj. \end{tabular}   & {$T_1$} & {$T_2$} & {$T_3$} \\

\midrule

3D-GS~\cite{kerbl20233d} & 13.81 & 11.70 & 13.50 \\
Scaffold-GS~\cite{lu2023scaffold} & 6.69 & 7.37 & 8.04 \\
Hierarchical-GS~\cite{kerbl2024hierarchical} & 9.13 & 8.54 & 8.91 \\
Hierarchical-GS($\tau_1$) & 16.14 & 13.26 & 14.79 \\
Hierarchical-GS($\tau_2$) & 19.70 & 19.59 & 18.94 \\
Hierarchical-GS($\tau_3$) & 24.33 & 25.29 & 24.75 \\
\midrule
Our-3D-GS & \textbf{57.08} & \textbf{56.85} & \textbf{56.07} \\
Our-Scaffold-GS & 40.91 & 35.17 & 40.31\\

\bottomrule
\end{tabular}
}
\end{table}

\begin{figure}[t!]
\centering
\includegraphics[width=\linewidth]{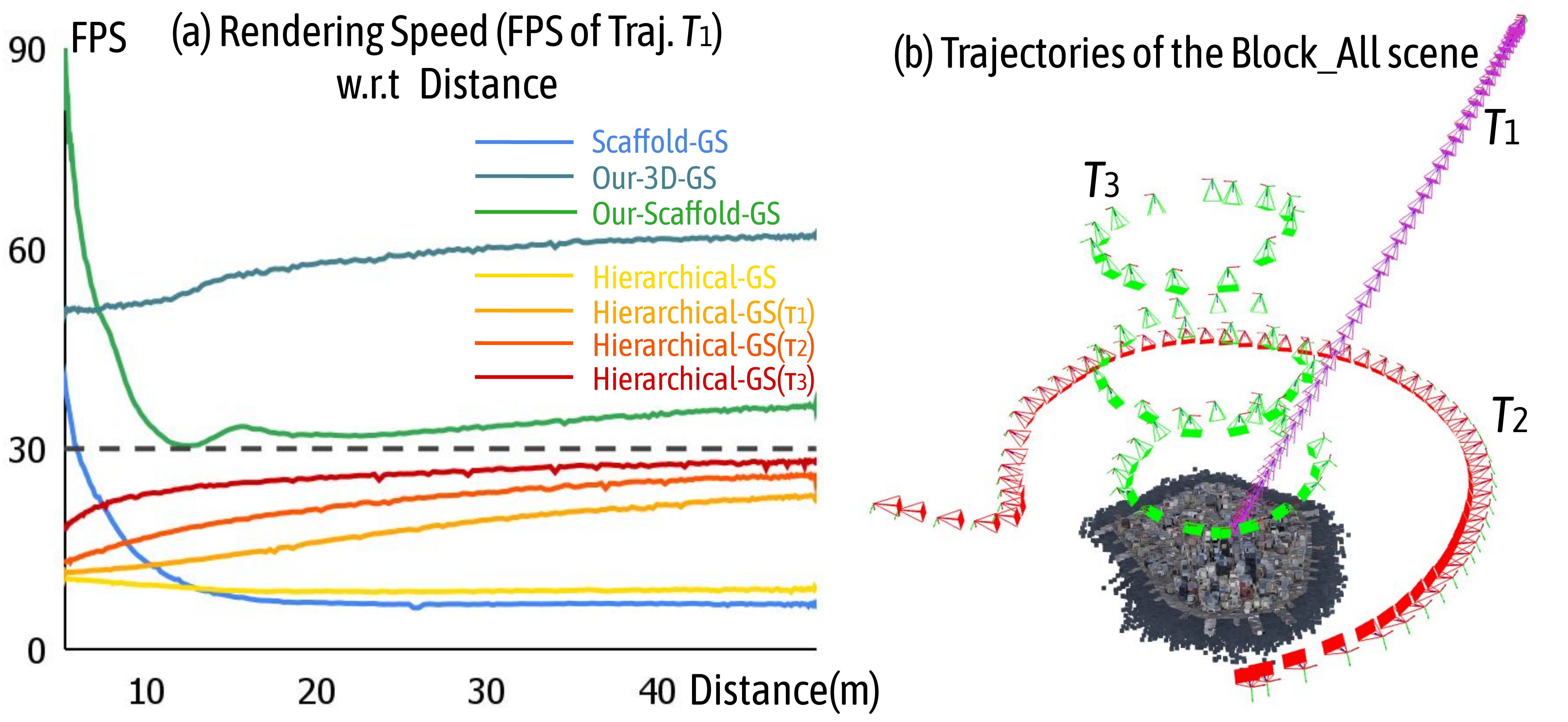}
\caption{
(a) The figure shows the rendering speed with respect to distance for different methods along trajectory $T_1$, both Our-3D-GS and Our-Scaffold-GS achieve real-time rendering speeds ($\geq 30$FPS). (b) The visualization depicts three different trajectories, corresponding to $T_1$, $T_2$, and $T_3$ in Tab.~\ref{tab:matrix_traj}, which are commonly found in video captures of large-scale scenes and illustrate the practical challenges involved.
}
\label{fig:traj}
\vspace{-1em}
\end{figure}

\subsubsection{Efficiency Analysis}

\paragraph{Rendering Time Comparisons}
Our goal is to enable real-time rendering of Gaussian representation models at any position within the scene using Level-of-Detail techniques. To evaluate our approach, we compare \modelname with three state-of-the-art methods~\cite{kerbl20233d, lu2023scaffold, kerbl2024hierarchical} on three novel view trajectories in Tab.~\ref{tab:matrix_traj} and Fig.~\ref{fig:traj}. These trajectories represent common movements in large-scale scenes, such as zoom-in, 360-degree circling, and multi-scale circling.
As shown in Tab.~\ref{tab:matrix_traj} and Fig.~\ref{fig:Rendering_comparison}, our method excels at capturing fine-grained details in close views while maintaining consistent rendering speeds at larger scales. Notably, our rendering speed is nearly $10 \times$ faster than Scaffold-GS~\cite{lu2023scaffold} in large-scale scenes and extreme-view sequences, which depends on our innovative LOD structure design.

\paragraph{Training Time Comparisons}
While our core contribution is the acceleration of rendering speed through LOD design, training speed is also critical for the practical application of photorealistic scene reconstruction. Below, we provide statistics for the Mip-NeRF360~\cite{barron2022mip} dataset (40k iterations): 2D-GS (28 mins), 3D-GS (34 mins), Mip-Splatting (46 mins), Scaffold-GS (29 mins), and Our-2D-GS (20 mins),  Our-3D-GS (21 mins), Our-Scaffold-GS (23 mins). 
Additionally, we report the training time for the concurrent work, Hierarchical-GS~\cite{kerbl2024hierarchical}. This method requires three stages to construct the LOD structure, which result in a longer training time (38 minutes for the first stage, totaling 69 minutes). In contrast, under the same number of iterations, our proposed method requires less time. Our-Scaffold-GS achieves the construction and optimization of the LOD structure in a single stage, taking only 35 minutes.
The reason our method can accelerate training time is twofold: the number of Gaussian primitives is relatively smaller, and not all Gaussians need to be optimized during progressive training.

\begin{table*}[htbp]
\centering
\renewcommand{\arraystretch}{1.1}
\setlength{\tabcolsep}{4pt}
\caption{Quantitative comparison on multi-resolution Mip-NeRF360~\cite{barron2022mip} dataset.
\modelname achieves better rendering quality across all scales compared to baselines.}
\label{tab:multi-reso}
\resizebox{1\linewidth}{!}{
\begin{tabular}{l|ccc|ccc|ccc|ccc|c}
\toprule
Scale Factor & \multicolumn{3}{c|}{$1 \times$} & \multicolumn{3}{c|}{$2 \times$} & \multicolumn{3}{c|}{$4 \times$} &\multicolumn{3}{c|}{$8 \times$} & \\
\begin{tabular}{c|c} Method & Metrics \end{tabular}  & PSNR\(\uparrow\) & SSIM\(\uparrow\) & LPIPS\(\downarrow\) & PSNR\(\uparrow\) & SSIM\(\uparrow\) & LPIPS\(\downarrow\) & PSNR\(\uparrow\) & SSIM\(\uparrow\) & LPIPS\(\downarrow\) & PSNR\(\uparrow\) & SSIM\(\uparrow\) & LPIPS\(\downarrow\) & \multirow{-2}{*}{\#GS(k)} \\
\midrule

3D-GS~\cite{kerbl20233d} & 26.16 & 0.757 & 0.301 & 27.33 & 0.822 & 0.202 & 28.55 & 0.884 & 0.117 & 27.85 & 0.897 & 0.086 & 430\\ 

Scaffold-GS~\cite{lu2023scaffold} & 26.81 & 0.767 & 0.285 & 28.09 & 0.835 & 0.183 & 29.52 & 0.898 & 0.099 & 28.98 & 0.915 & 0.072 & 369\\ 

Mip-Splatting~\cite{yu2023mip} & \underline{27.43} & \textbf{0.801} & \textbf{0.244} & \underline{28.56} & \textbf{0.857} & \textbf{0.152} & \underline{30.00} & \textbf{0.910} & \underline{0.087} & \underline{31.05} & \textbf{0.942} & \textbf{0.055} & 642\\ [0.5ex]

\hline

Our-Scaffold-GS & \textbf{27.68} & \underline{0.791} & \underline{0.245} & \textbf{28.82} & \underline{0.850} & \underline{0.157} & \textbf{30.27} & \underline{0.906} & \textbf{0.087} & \textbf{31.18} & \underline{0.932} & \underline{0.057} & 471\\

\bottomrule
\end{tabular}}
\end{table*}

\begin{figure*}[t!]
\centering
\includegraphics[width=\linewidth]{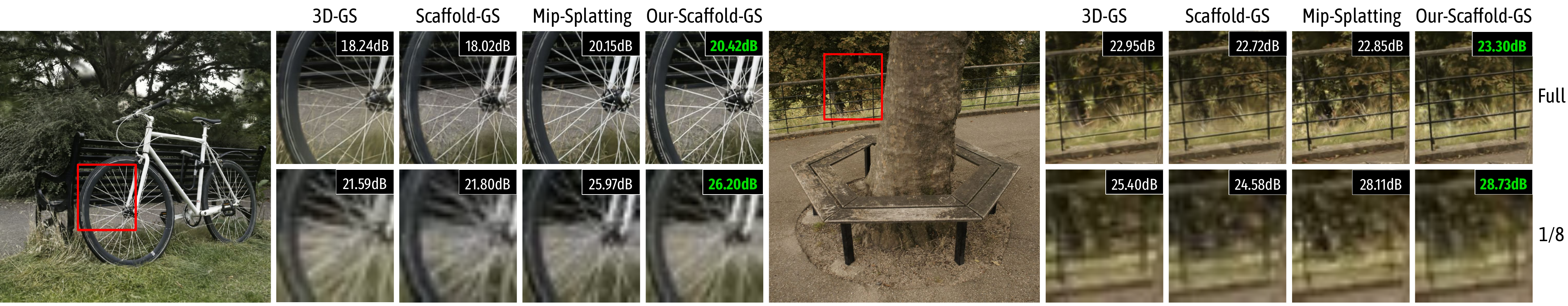}
\caption{Qualitative comparison of full-resolution and low-resolution (1/8 of full-resolution) on multi-resolution Mip-NeRF360\cite{barron2022mip} datasets. Our approach demonstrates adaptive anti-aliasing and effectively recovers fine-grained details, while baselines often produce artifacts, particularly on elongated structures such as bicycle wheels and handrails.
}
\label{fig:multiresolution}
\vspace{-1em}
\end{figure*}

\subsubsection{Robustness Analysis}

\begin{figure}[t!]
\centering
\includegraphics[width=\linewidth]{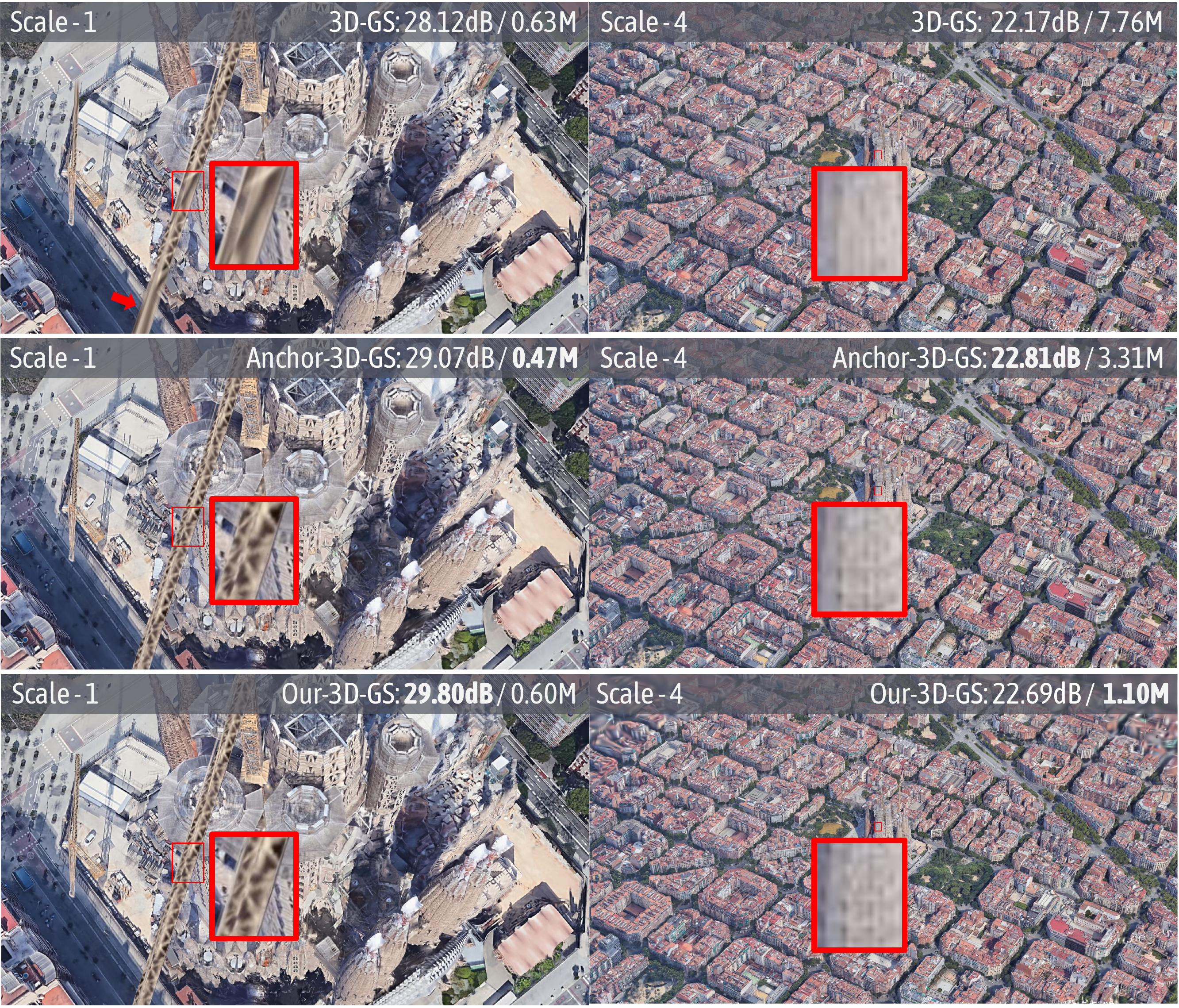}
\caption{Qualitative comparison of scale-1 and scale-4 on the Barcelona scene from the BungeeNeRF~\cite{xiangli2022bungeenerf} dataset. Both Anchor-3D-GS and Our-3D-GS accurately reconstruct fine details, such as the crane in scale-1 and the building surface in scale-4 (see highlighted patches and arrows), while Our-3D-GS uses fewer primitives to model the entire scene. We report PSNR and the number of Gaussians used for rendering.
}
\label{fig:multiscale}
\vspace{-0.8em}
\end{figure}

\paragraph{Multi-Scale Results}
To evaluate the ability of \modelname to handle multi-scale scene details, we conduct an experiment using the BungeeNeRF~\cite{xiangli2022bungeenerf} dataset across four different scales (\ie, from ground-level to satellite-level camera altitudes). Our results show that \modelname accurately captures scene details and models the entire scene more efficiently with fewer Gaussian primitives, as demonstrated in Tab.~\ref{tab:bungeenerf} and Fig.~\ref{fig:multiscale}.

\paragraph{Multi-Resolution Results}
\label{sec: Multi-Resolution_Results}
As mentioned in Sec.~\ref{sec:method}, when dealing with training views that vary in camera resolution or intrinsics, such as datasets presented in~\cite{barron2022mip} with a four-fold downsampling operation, 
we multiply the observation distance with factor scale factor accordingly to handle this multi-resolution dataset.
As shown in Fig.~\ref{fig:multiresolution} and Tab. \ref{tab:multi-reso}, we train all models on images with downsampling scales of 1, 2, 4, 8, and \modelname adaptively handle the changed footprint size and effectively address the aliasing issues inherent to 3D-GS~\cite{kerbl20233d} and Scaffold-GS~\cite{lu2023scaffold}. As resolution changes, 3D-GS and Scaffold-GS introduce noticeable erosion artifacts, but our approach avoids such issues, achieving results competitive with Mip-Splatting~\cite{yu2023mip} and even closer to the ground truth.
Additionally, we provide multi-resolution results for the Tanks\&Temples dataset~\cite{knapitsch2017tanks} and the Deep Blending dataset~\cite{hedman2018deep} in the supplementary materials.

\paragraph{Random Initialization Results}
To illustrate the independence of our framework from SfM points, we evaluate it using randomly initialized points, with 0.31/0.27 (LPIPS$\downarrow$), 25.93/26.41 (PSNR$\uparrow$), 0.76/0.77 (SSIM$\uparrow$) on Mip-NeRF360~\cite{barron2022mip} dataset comparing Scaffold-GS with Our-Scaffold-GS. 
The improvement primarily depends on the efficient densification strategy.

\begin{figure}[t!]
\centering
\includegraphics[width=\linewidth]{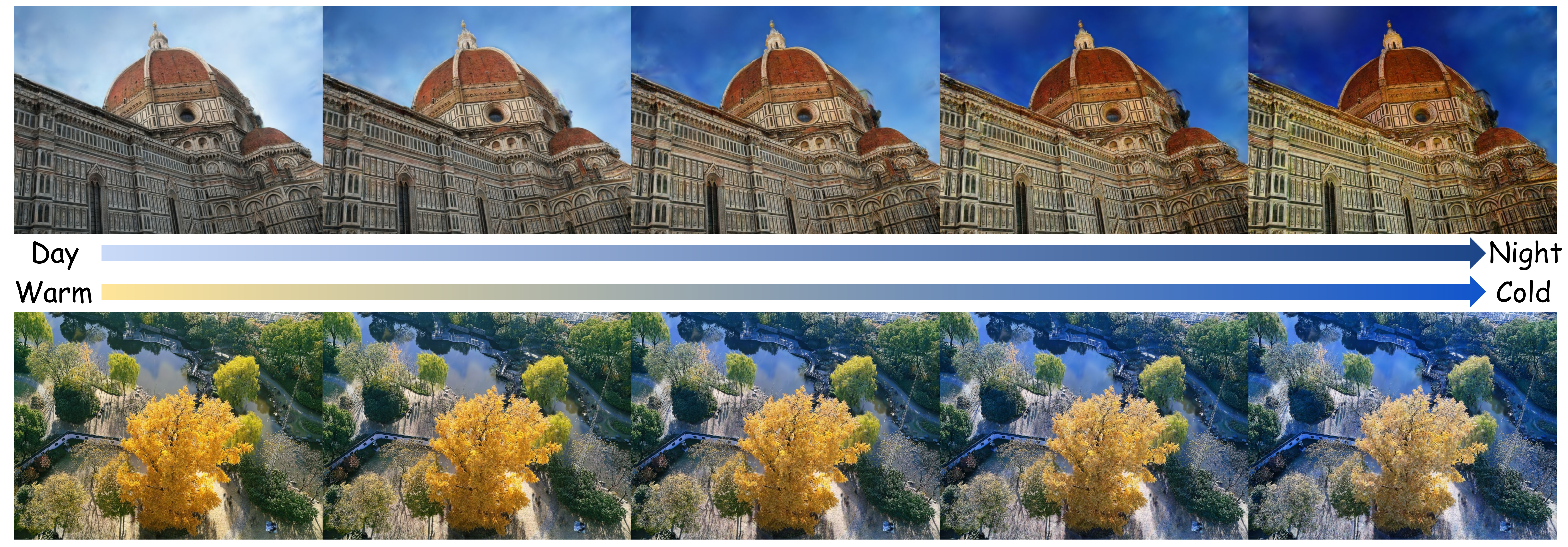}
\caption{Visualization of appearance code interpolation. We show five test views from the Phototourism~\cite{jin2021image} dataset ~(top) and a self-captured tree scene (bottom) with linearly-interpolated appearance codes.
}
\label{fig:ape}
\vspace{-1em}
\end{figure}

\paragraph{Appearance Embedding Results}
We demonstrate that our specialized design can handle input images with different exposure compensations and provide detailed control over lighting and appearance. As shown in Fig.~\ref{fig:ape}, we reconstruct two scenes: one is from the widely-used Phototourism~\cite{snavely2006photo} dataset and the other is a self-captured scene of a ginkgo tree. We present five images rendered from a fixed camera view, where we interpolate the appearance codes linearly to produce a fancy style transfer effect.

\subsection{Ablation Studies}
\label{sec:ablation}

In this section, we ablate each individual module to validate their effectiveness. 
We select all scenes from the Mip-NeRF360~\cite{barron2022mip} dataset as quantitative comparison, given its representative characteristics. 
Additionally, we select Block\_Small from the MatrixCity~\cite{li2023matrixcity} dataset for qualitative comparison.
In this section, we ablate each individual module to verify their effectiveness. Meanwhile, we choose the octree-version of Scaffold-GS as the full model, with the vanilla Scaffold-GS serving as the baseline for comparison.
Quantitative and qualitative results can be found in Tab.~\ref{tab:ablation} and Fig.~\ref{fig:ablate}.

\subsubsection{Next Level Grow Operator} 

To evaluate the effectiveness of next-level anchor growing, as detailed in Section \ref{method: anchor_refinement}, we conduct an ablation in which new anchors are only allowed to grow at the same LOD level. The results, presented in Tab. \ref{tab:ablation}, show that while the number of rendered Gaussian primitives and storage requirements decreased, there was a significant decline in image visual quality. This suggests that incorporating finer anchors into higher LOD levels not only improves the capture of high-frequency details but also enhances the interaction between adjacent LOD levels.

\subsubsection{LOD Bias}
To validate its contribution to margin details, we ablate the proposed LOD bias. The results, presented in Tab.~\ref{tab:ablation}, indicates that LOD bias is essential for enhancing the rendering quality, particularly in regions rich in high-frequency details for smooth trajectories, which can be observed in column (a)(b) of Fig.~\ref{fig:ablate}, as the white stripes on the black buildings become continuous and complete.


\subsubsection{Progressive Training}

\begin{figure}[t!]
\centering
\includegraphics[width=\linewidth]{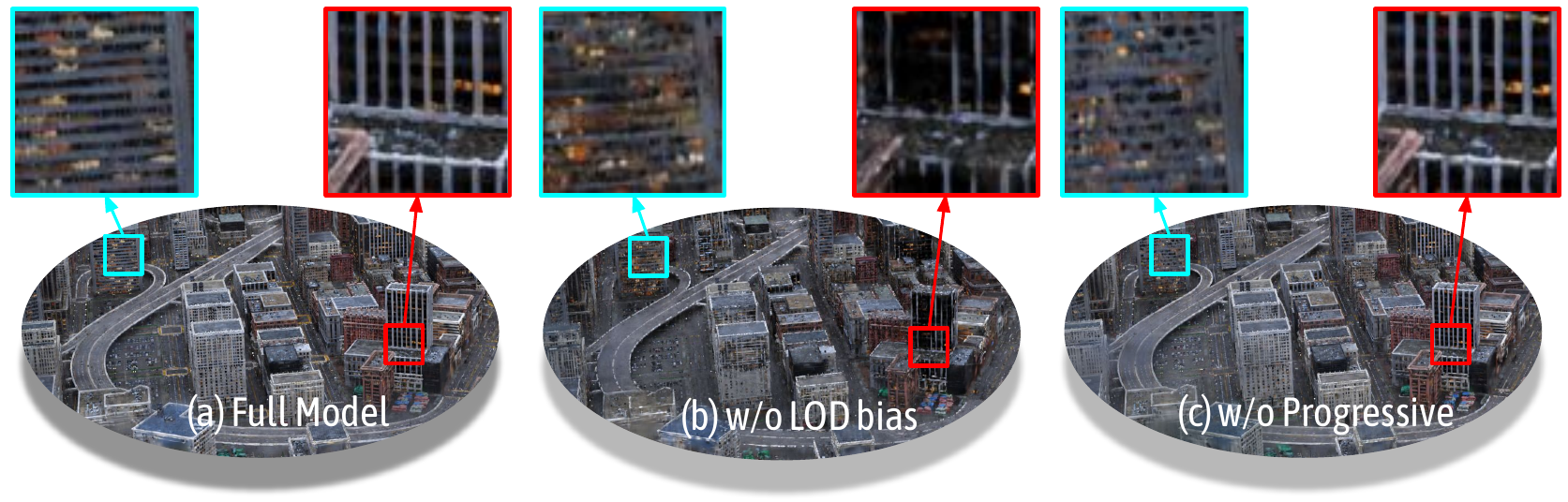}
\caption{Visualizations of the rendered images from (a) our full model, (b) ours w/o LOD bias, (c) ours w/o progressive training. As observed, LOD bias aids in restoring sharp building edges and lines, while progressive training helps recover the geometric structure from coarse to fine details.
}
\label{fig:ablate}
\centering
\vspace{-1em}
\end{figure}

To compare its influence on LOD level overlapping, we ablate progressive training strategy. In column (a)(c) of Fig. \ref{fig:ablate}, the building windows are clearly noticeable, indicating that the strategy contributes to reduce the rendered Gaussian redundancy and decouple the Gaussias of different scales in the scene to their corresponding LOD levels. In addition, the quantitative results also verify the improvement of scene reconstruction accuracy by the proposed strategy, as shown in Tab.~\ref{tab:ablation}.

\subsubsection{View Frequency}
Due to the design of the octree structure, anchors at higher LOD levels are only rendered and optimized when the camera view is close to them. These anchors are often not sufficiently optimized due to their limited number, leading to visual artifacts when rendering from novel views.
We perform an ablation of the view frequency strategy during the anchor pruning stage, as described detailly in Sec.~\ref{method:anchor_prune}. Implementing this strategy eliminates floaters, particularly in close-up views, enhances visual quality, and significantly reduces storage requirements, as shown in Tab.~\ref{tab:ablation} and Fig.~\ref{fig:view_frequency}.

\begin{table}[t!]
\centering
\renewcommand{\arraystretch}{1.15}
\setlength{\tabcolsep}{3pt}
\caption{Quantitative results on ablation studies. We list the rendering metrics for each ablation described in Sec.~\ref{sec:ablation}.
}
\vspace{-6pt}
\label{tab:ablation}
\resizebox{\linewidth}{!}{
\begin{tabular}{l|cccc}
\toprule
Method & PSNR($\uparrow$) & SSIM($\uparrow$) & LPIPS($\downarrow$) & \#GS(k)/Mem \\

\midrule

Scaffold-GS~\cite{lu2023scaffold} & \underline{27.90} & 0.815 & 0.220 & 666/197.5M \\

Ours w/o $l_{next}$ grow. & 27.64 & 0.811 & 0.223 & \textbf{594/99.7M} \\

Ours w/o progressive. & 27.86 & 0.818 & 0.215 & 698/142.3M \\

Ours w/o LOD bias & 27.85 & \underline{0.818} & 0.214 & 667/146.8M \\

Ours w/o view freq. & 27.74 & 0.817 & \textbf{0.211} & 765/244.4M \\

Our-Scaffold-GS & \textbf{28.05} & \textbf{0.819} & \underline{0.214} & \underline{657/139.6M} \\

\bottomrule
\end{tabular}}
\end{table}

\section{Limitations and Conclusion}
\label{sec:conclusion}
In this work, we introduce Level-of-Details (LOD) to Gaussian representation, using a novel octree structure to organize anchors hierarchically. Our model, \modelname, addresses previous limitations by dynamically fetching appropriate LOD levels based on observed views and scene complexity, ensuring consistent rendering performance with adaptive LOD adjustments. Through careful design, \modelname significantly enhances detail capture while maintaining real-time rendering performance without increasing the number of Gaussian primitives. This suggests potential for future real-world streaming experiences, demonstrating the capability of advanced rendering methods to deliver seamless, high-quality interactive 3D scene and content.

However, certain model components, like octree construction and progressive training, still require hyperparameter tuning.
Balancing anchors in each LOD level and adjusting training iteration activation are also crucial. 
Moreover, our model still faces challenges associated with 3D-GS, including dependency on the precise camera poses and lack of geometry support. These are left as our future works.

\vfill

\newpage
\section{Supplementary Material}
The supplementary material includes quantitative results for each scene from the dataset used in the main text, covering image quality metrics such as PSNR, ~\cite{wang2004image} and LPIPS~\cite{zhang2018unreasonable}, as well as the number of rendered Gaussian primitives and storage size.

\begin{table}[htbp]
\renewcommand{\arraystretch}{1.1}
\setlength{\tabcolsep}{1pt}
\centering
\caption{PSNR for all scenes in the Mip-NeRF360\cite{barron2022mip} dataset.}
\vspace{-6pt}
\resizebox{\linewidth}{!}{
\begin{tabular}{l|ccccccccc}
\toprule
\begin{tabular}{c|c} Method & Scenes \end{tabular} & bicycle & bonsai &  counter & flowers & garden & kitchen & room & stump & treehill \\
\midrule

2D-GS\cite{huang20242d} & 24.77 & 31.42 & 28.20 & 21.02 & 26.73 & 30.66 & 30.95 & 26.17 & 22.48 \\
3D-GS\cite{kerbl20233d} & 25.10 & 32.19 & 29.22 & 21.57 & 27.45 & 31.62 & 31.53 & 26.70 & 22.46 \\
Mip-Splatting\cite{yu2023mip} & 25.13 & 32.56 & 29.30 & 21.64 & 27.43 & 31.48 & 31.73 & 26.65 & 22.60 \\
Scaffold-GS\cite{lu2023scaffold} & 25.19 & 33.22 & 29.99 & 21.40 & 27.48 & 31.77 & 32.30 & 26.67 & 23.08 \\
\midrule
Anchor-2D-GS & 24.81 & 31.01 & 28.44 & 21.25 & 26.65 & 30.35 & 31.08 & 26.52 & 22.72 \\
Anchor-3D-GS & 25.21 & 32.20 & 29.12 & \textbf{21.52} & 27.37 & 31.46 & 31.83 & 26.74 & 22.85 \\
\midrule
Our-2D-GS & 24.89 & 30.85 & 28.56 & 21.19 & 26.88 & 30.22 & 31.17 & 26.62 & 22.78 \\
Our-3D-GS & 25.20 & 32.29 & 29.27 & 21.40 & 27.36 & 31.70 & 31.96 & \textbf{26.78} & 22.85 \\
Our-Scaffold-GS & \textbf{25.24} & \textbf{33.76} & \textbf{30.19} & 21.46 & \textbf{27.67} & \textbf{31.84} & \textbf{32.51} & 26.63 & \textbf{23.13} \\
\bottomrule

\end{tabular}
}
\end{table}

\vspace{-0.5em}
\begin{table}[htbp]
\renewcommand{\arraystretch}{1.1}
\setlength{\tabcolsep}{1pt}
\centering
\caption{SSIM for all scenes in the Mip-NeRF360\cite{barron2022mip} dataset.}
\vspace{-6pt}
\resizebox{\linewidth}{!}{
\begin{tabular}{l|ccccccccc}
\toprule
\begin{tabular}{c|c} Method & Scenes \end{tabular} & bicycle & bonsai &  counter & flowers & garden & kitchen & room & stump & treehill \\
\midrule

2D-GS\cite{huang20242d} & 0.730 & 0.935 & 0.899 & 0.568 & 0.839 & 0.923 & 0.916 & 0.759 & 0.627 \\
3D-GS\cite{kerbl20233d} & 0.747 & 0.947 & 0.917 & 0.600 & 0.861 & 0.932 & 0.926 & \textbf{0.773} & 0.636 \\
Mip-Splatting\cite{yu2023mip} & 0.747 & 0.948 & 0.917 & \textbf{0.601} & 0.861 & 0.933 & 0.928 & 0.772 & 0.639 \\
Scaffold-GS\cite{lu2023scaffold} & 0.751 & 0.952 & 0.922 & 0.587 & 0.853 & 0.931 & 0.932 & 0.767 & \textbf{0.644} \\
\midrule
Anchor-2D-GS & 0.735 & 0.933 & 0.900 & 0.575 & 0.838 & 0.917 & 0.917 & 0.762 & 0.630 \\
Anchor-3D-GS & 0.758 & 0.946 & 0.913 & 0.591 & 0.857 & 0.928 & 0.927 & 0.772 & 0.640 \\
\midrule
Our-2D-GS & 0.737 & 0.932 & 0.903 & 0.572 & 0.838 & 0.918 & 0.919 & 0.763 & 0.630 \\
Our-3D-GS & \textbf{0.761} & 0.946 & 0.916 & 0.587 & 0.855 & 0.931 & 0.929 & 0.772 & 0.640 \\
Our-Scaffold-GS & 0.755 & \textbf{0.955} & \textbf{0.925} & 0.595 & \textbf{0.861} & \textbf{0.933} & \textbf{0.936} & 0.766 & 0.641 \\
\bottomrule

\end{tabular}
}
\end{table}

\vspace{-0.5em}
\begin{table}[htbp]
\renewcommand{\arraystretch}{1.1}
\setlength{\tabcolsep}{1pt}
\centering
\caption{LPIPS for all scenes in the Mip-NeRF360\cite{barron2022mip} dataset.}
\vspace{-6pt}
\resizebox{\linewidth}{!}{
\begin{tabular}{l|ccccccccc}
\toprule
\begin{tabular}{c|c} Method & Scenes \end{tabular} & bicycle & bonsai &  counter & flowers & garden & kitchen & room & stump & treehill \\
\midrule

2D-GS\cite{huang20242d} & 0.284 & 0.204 & 0.214 & 0.389 & 0.153 & 0.134 & 0.218 & 0.279 & 0.385 \\
3D-GS\cite{kerbl20233d} & 0.243 & 0.178 & 0.179 & 0.345 & 0.114 & 0.117 & 0.196 & 0.231 & 0.335 \\
Mip-Splatting\cite{yu2023mip} & 0.245 & 0.178 & 0.179 & 0.347 & 0.115 & 0.115 & 0.192 & 0.232 & \textbf{0.334} \\
Scaffold-GS\cite{lu2023scaffold} & 0.247 & 0.173 & 0.177 & 0.359 & 0.13 & 0.118 & 0.183 & 0.252 & 0.338 \\
\midrule
Anchor-2D-GS & 0.262 & 0.200 & 0.203 & 0.376 & 0.146 & 0.140 & 0.209 & 0.261 & 0.371 \\
Anchor-3D-GS & 0.230 & 0.177 & 0.182 & 0.363 & 0.121 & 0.121 & 0.193 & \textbf{0.249} & 0.348 \\
\midrule
Our-2D-GS & 0.262 & 0.205 & 0.198 & 0.378 & 0.148 & 0.140 & 0.205 & 0.264 & 0.374 \\
Our-3D-GS & \textbf{0.225} & 0.178 & 0.176 & 0.364 & 0.125 & 0.116 & 0.190 & 0.250 & 0.357
\\
Our-Scaffold-GS & 0.235 & \textbf{0.164} & \textbf{0.169} & \textbf{0.347} & \textbf{0.116} & \textbf{0.115} & \textbf{0.172} & 0.250 & 0.360 \\
\bottomrule

\end{tabular}
}
\end{table}

\vspace{-0.5em}
\begin{table}[htbp]
\renewcommand{\arraystretch}{1.1}
\setlength{\tabcolsep}{1pt}
\centering
\caption{Number of Gaussian Primitives(\#K) for all scenes in the Mip-NeRF360\cite{barron2022mip} dataset.}
\vspace{-6pt}

\resizebox{\linewidth}{!}{
\begin{tabular}{l|ccccccccc}
\toprule
\begin{tabular}{c|c} Method & Scenes \end{tabular} & bicycle & bonsai &  counter & flowers & garden & kitchen & room & stump & treehill \\
\midrule

2D-GS\cite{huang20242d} & 555 & 210 & 232 & 390 & 749 & 440 & 199 & 413 & 383 \\
3D-GS\cite{kerbl20233d} & 1453 & 402 & 530 & 907 & 2030 & 1034 & 358 & 932 & 785 \\
Mip-Splatting\cite{yu2023mip} & 1584 & 430 & 545 & 950 & 2089 & 1142 & 405 & 1077 & 892 \\
Scaffold-GS\cite{lu2023scaffold} & 764 & 532 & 377 & 656 & 1121 & 905 & 272 & 637 & 731 \\
\midrule
Anchor-2D-GS & 887 & 337 & 353 & 548 & 938 & 466 & 270 & 587 & 540 \\
Anchor-3D-GS & 1187 & 370 & 388 & 634 & 1524 & 535 & 293 & 647 & 781 \\
\midrule
Our-2D-GS & \textbf{540} & \textbf{259} & \textbf{294} & \textbf{428} & \textbf{718} & \textbf{414} & \textbf{184} & \textbf{394} & \textbf{344} \\
Our-3D-GS & 659 & 301 & 334 & 478 & 987 & 710 & 195 & 436 & 433 \\
Our-Scaffold-GS & 653 & 631 & 409 & 675 & 1475 & 777 & 374 & 549 & 372 \\
\bottomrule

\end{tabular}
}
\end{table}

\vspace{-2em}
\begin{table}[htbp]
\renewcommand{\arraystretch}{1.1}
\setlength{\tabcolsep}{1pt}
\centering
\caption{Storage memory(\#MB) for all scenes in the Mip-NeRF360\cite{barron2022mip} dataset.}
\vspace{-6pt}

\resizebox{\linewidth}{!}{
\begin{tabular}{l|ccccccccc}
\toprule
\begin{tabular}{c|c} Method & Scenes \end{tabular} & bicycle & bonsai &  counter & flowers & garden & kitchen & room & stump & treehill \\
\midrule

2D-GS\cite{huang20242d} & 889.6 & 173.1 & 135.4 & 493.5 & 603.1 & 191.0 & 180.0 & 670.3 & 630.9\\
3D-GS\cite{kerbl20233d} & 1361.8 & 293.5 & 293.3 & 878.5 & 1490.6 & 413.1 & 355.6 & 1115.2 & 878.6 \\
Mip-Splatting\cite{yu2023mip} & 1433.6 & 318.1 & 307.5 & 970.2 & 1448.9 & 463.4 & 401.0 & 1239.0 & 964.3 \\
Scaffold-GS\cite{lu2023scaffold} & 340.2 & 133.3 & 90.4 & 243.8 & 231.7 & 102.2 & 86.1 & 294.2 & 256.0 \\
\midrule
Anchor-2D-GS & 599.2 & 280.0 & 191.5 & 530.0 & 634.4 & 190.7 & 228.4 & 359.1 & 521.4 \\
Anchor-3D-GS & 765.5 & 301.7 & 204.9 & 656.1 & 988.6 & 217.0 & 244.6 & 417.4 & 632.2 \\
\midrule
Our-2D-GS & 485.0 & 368.6 & 265.6 & 442.3 & 598.6 & 272.3 & 180.8 & 292.8 & 438.3 \\
Our-3D-GS & 648.6 & 382.7 & 305.8 & 487.7 & 706.2 & 282.9 & 162.7 & 322.1 & 468.4 \\
Our-Scaffold-GS & \textbf{216.0} & \textbf{133.5} & \textbf{83.2} & \textbf{198.3} & \textbf{236.3} & \textbf{88.7} & \textbf{83.5} & \textbf{141.9} & \textbf{104.4} \\
\bottomrule

\end{tabular}
}
\end{table}

\vspace{-2em}
\begin{table}[htbp]
\centering
\renewcommand{\arraystretch}{1.05}
\setlength{\tabcolsep}{1pt}
\caption{Quantitative results for all scenes in the Tanks\&Temples\cite{knapitsch2017tanks} dataset.}
\vspace{-6pt}
\resizebox{\linewidth}{!}{
\begin{tabular}{l|cccc|cccc}
\toprule
Dataset & \multicolumn{4}{c|}{Truck} & \multicolumn{4}{c}{Train} \\

\begin{tabular}{c|c} Method & Metrics \end{tabular}  & PSNR  & SSIM  & LPIPS  & \#GS(k)/Mem & PSNR  & SSIM  & LPIPS  & \#GS(k)/Mem \\
\midrule
2D-GS\cite{huang20242d} & 25.12 & 0.870 & 0.173 & 393/287.2M & 21.38 & 0.790 & 0.251 & 310/121.5M \\
3D-GS\cite{kerbl20233d} & 25.52 & 0.884 & 0.142 & 876/610.8M & 22.30 & 0.819 & 0.201 & 653/249.3M \\
Mip-Splatting\cite{yu2023mip} & 25.74 & 0.888 & 0.142 & 967/718.9M & 22.17 & 0.824 & 0.199 & 696/281.9M \\
Scaffold-GS\cite{lu2023scaffold} & 26.04 & 0.889 & 0.131 & 698/214.6M & 22.91 & 0.838 & 0.181 & 554/120.4M \\
\midrule
Anchor-2D-GS & 25.45 & 0.873 & 0.161 & 472/349.7M & 21.58 & 0.797 & 0.237 & 457/208.3M \\
Anchor-3D-GS & 25.85 & 0.883 & 0.146 & 603/452.8M & 22.18 & 0.810 & 0.222 & 541/245.6M \\
\midrule
Our-2D-GS & 25.32 & 0.872 & 0.158 & \textbf{304}/208.5M & 21.92 & 0.812 & 0.215 & \textbf{355}/173.9M \\
Our-3D-GS & 25.81 & 0.887 & 0.131 & 407/542.8M & 22.52 & 0.828 & 0.190 & 440/224.90M \\
Our-Scaffold-GS & \textbf{26.24} & \textbf{0.894} & \textbf{0.122} & 426/\textbf{93.7M} & \textbf{23.11} & \textbf{0.838} & \textbf{0.184} & 460/\textbf{83.4M} \\
\bottomrule

\end{tabular}
}
\end{table}

\vspace{-2em}
\begin{table}[htbp]
\centering
\renewcommand{\arraystretch}{1.05}
\setlength{\tabcolsep}{1pt}
\caption{Quantitative results for all scenes in the DeepBlending\cite{hedman2018deep} dataset.}
\vspace{-6pt}

\resizebox{\linewidth}{!}{
\begin{tabular}{l|cccc|cccc}
\toprule
Dataset & \multicolumn{4}{c|}{Dr Johnson} & \multicolumn{4}{c}{Playroom} \\

\begin{tabular}{c|c} Method & Metrics \end{tabular}  & PSNR  & SSIM  & LPIPS  & \#GS(k)/Mem & PSNR  & SSIM  & LPIPS  & \#GS(k)/Mem \\
\midrule
2D-GS\cite{huang20242d} & 28.74 & 0.897 & 0.257 & 232/393.8M & 29.89 & 0.900 & 0.257 & 160/276.7M \\
3D-GS\cite{kerbl20233d} & 29.09 & 0.900 & 0.242 & 472/818.9M & 29.83 & 0.905 & 0.241 & 324/592.3M \\
Mip-Splatting\cite{yu2023mip} & 29.08 & 0.900 & 0.241 & 512/911.6M
 & 30.03 & 0.902 & 0.245 & 307/562.0M \\
Scaffold-GS\cite{lu2023scaffold} & 29.73 & \textbf{0.910} & \textbf{0.235} & 232/145.0M & 30.83 & 0.907 & \textbf{0.242} & 182/106.0M \\
\midrule
Anchor-2D-GS & 28.68 & 0.893 & 0.266 & 186/346.3M & 30.02 & 0.899 & 0.262 & 138/231.8M \\
Anchor-3D-GS & 29.23 & 0.897 & 0.267 & 141/242.3M & 30.08 & 0.901 & 0.252 & 159/303.4M \\
\midrule
Our-2D-GS & 28.94 & 0.894 & 0.26 & 97/268.2M & 29.93 & 0.899 & 0.268 & 70/136.4M \\
Our-3D-GS & 29.27 & 0.900 & 0.251 & \textbf{95}/240.7M & 30.03 & 0.901 & 0.263 & \textbf{63}/119.2M \\
Our-Scaffold-GS & \textbf{29.83} & 0.909 & 0.237 & 124/\textbf{92.46M} & \textbf{31.15} & \textbf{0.914} & 0.245 & 100/\textbf{50.91M} \\
\bottomrule

\end{tabular}
}
\end{table}

\vspace{-2em}
\begin{table}[htbp]
\renewcommand{\arraystretch}{1.1}
\setlength{\tabcolsep}{1pt}
\centering
\caption{PSNR for all scenes in the BungeeNeRF\cite{xiangli2022bungeenerf} dataset.}
\vspace{-6pt}

\resizebox{\linewidth}{!}{
\begin{tabular}{l|ccccccccc}
\toprule
\begin{tabular}{c|c} Method & Scenes \end{tabular}  & Amsterdam & Barcelona & Bilbao & Chicago & Hollywood & Pompidou & Quebec & Rome\\
\midrule

2D-GS\cite{huang20242d} & 27.22 & 27.01 & 28.59 & 25.62 & 26.43 & 26.62 & 28.38 & 26.95 \\
3D-GS\cite{kerbl20233d} & 27.75 & 27.55 & 28.91 & 28.27 & 26.25 & 27.16 & 28.86 & 27.56 \\
Mip-Splatting\cite{yu2023mip} & \textbf{28.16} & 27.72 & 29.13 & 28.28 & 26.59 & 27.71 & 29.23 & 28.33 \\
Scaffold-GS\cite{lu2023scaffold} & 27.82 & 28.09 & 29.20 & 28.55 & 26.36 & \textbf{27.72} & 29.29 & 28.24 \\
\midrule
Anchor-2D-GS & 26.80 & 27.03 & 28.02 & 27.50 & 25.68 & 26.87 & 28.21 & 27.32 \\
Anchor-3D-GS & 27.70 & 27.93 & 28.92 & 28.20 & 26.20 & 27.17 & 28.83 & 28.22 \\
\midrule
Our-2D-GS & 27.14 & 27.28 & 28.24 & 27.78 & 26.13 & 26.58 & 28.07 & 27.47 \\
Our-3D-GS & 27.95 & 27.91 & 28.81 & 28.24 & 26.51 & 27.00 & 28.98 & 28.09 \\
Our-Scaffold-GS & 28.16 & \textbf{28.40} & \textbf{29.39} & \textbf{28.86} & \textbf{26.76} & 27.46 & \textbf{29.46} & \textbf{28.59} \\
\bottomrule

\end{tabular}
}
\end{table}

\vspace{-2em}
\begin{table}[htbp]
\renewcommand{\arraystretch}{1.1}
\setlength{\tabcolsep}{1pt}
\centering
\caption{SSIM for all scenes in the BungeeNeRF\cite{xiangli2022bungeenerf} dataset.}
\vspace{-6pt}

\resizebox{\linewidth}{!}{
\begin{tabular}{l|ccccccccc}
\toprule
\begin{tabular}{c|c} Method & Scenes \end{tabular} & Amsterdam & Barcelona & Bilbao & Chicago & Hollywood & Pompidou & Quebec & Rome\\
\midrule

2D-GS\cite{huang20242d} & 0.896 & 0.907 & 0.912 & 0.901 & 0.872 & 0.907 & 0.923 & 0.902 \\
3D-GS\cite{kerbl20233d} & 0.918 & 0.919 & 0.918 & 0.932 & 0.873 & 0.919 & 0.937 & 0.918 \\
Mip-Splatting\cite{yu2023mip} & 0.918 & 0.919 & 0.918 & 0.930 & 0.876 & 0.923 & 0.938 & 0.922 \\
Scaffold-GS\cite{lu2023scaffold} & 0.914 & 0.923 & 0.918 & 0.929 & 0.866 & \textbf{0.926} & 0.939 & 0.924 \\
\midrule
Anchor-2D-GS & 0.872 & 0.887 & 0.886 & 0.897 & 0.838 & 0.900 & 0.910 & 0.891 \\
Anchor-3D-GS & 0.902 & 0.912 & 0.907 & 0.916 & 0.871 & 0.919 & 0.930 & 0.915 \\
\midrule
Our-2D-GS & 0.887 & 0.894 & 0.892 & 0.912 & 0.857 & 0.893 & 0.911 & 0.895 \\
Our-3D-GS & 0.912 & 0.910 & 0.905 & 0.920 & 0.875 & 0.907 & 0.928 & 0.912 \\
Our-Scaffold-GS & \textbf{0.922} & \textbf{0.928} & \textbf{0.921} & \textbf{0.934} & \textbf{0.884} & 0.923 & \textbf{0.942} & \textbf{0.930} \\
\bottomrule

\end{tabular}
}
\end{table}

\vspace{-2em}
\begin{table}[htbp]
\renewcommand{\arraystretch}{1.1}
\setlength{\tabcolsep}{1pt}
\centering
\caption{LPIPS for all scenes in the BungeeNeRF\cite{xiangli2022bungeenerf} dataset.}
\vspace{-6pt}

\resizebox{\linewidth}{!}{
\begin{tabular}{l|ccccccccc}
\toprule
\begin{tabular}{c|c} Method & Scenes \end{tabular} & Amsterdam & Barcelona & Bilbao & Chicago & Hollywood & Pompidou & Quebec & Rome\\
\midrule

2D-GS\cite{huang20242d} & 0.132 & 0.101 & 0.109 & 0.13 & 0.152 & 0.109 & 0.113 & 0.123 \\
3D-GS\cite{kerbl20233d} & 0.092 & 0.082 & 0.092 & 0.080 & 0.128 & 0.090 & 0.087 & 0.096 \\
Mip-Splatting\cite{yu2023mip} & 0.094 & 0.082 & 0.095 & 0.081 & 0.130 & 0.087 & 0.087 & 0.093 \\
Scaffold-GS\cite{lu2023scaffold} & 0.102 & 0.078 & \textbf{0.090} & 0.08 & 0.157 & \textbf{0.082} & \textbf{0.080} & 0.087 \\
\midrule
Anchor-2D-GS & 0.156 & 0.125 & 0.137 & 0.125 & 0.196 & 0.119 & 0.127 & 0.131 \\
Anchor-3D-GS & 0.127 & 0.099 & 0.119 & 0.105 & 0.160 & 0.100 & 0.100 & 0.105 \\
\midrule
Our-2D-GS & 0.139 & 0.112 & 0.131 & 0.103 & 0.169 & 0.126 & 0.125 & 0.128 \\
Our-3D-GS & 0.105 & 0.094 & 0.115 & 0.095 & 0.146 & 0.113 & 0.100 & 0.108 \\
Our-Scaffold-GS & \textbf{0.090} & \textbf{0.071} & 0.091 & \textbf{0.077} & \textbf{0.128} & 0.089 & 0.081 & \textbf{0.080} \\
\bottomrule

\end{tabular}
}
\end{table}

\vspace{-0.5em}
\begin{table}[htbp]
\renewcommand{\arraystretch}{1.1}
\setlength{\tabcolsep}{1pt}
\centering
\caption{Number of Gaussian Primitives(\#K) for all scenes in the BungeeNeRF\cite{xiangli2022bungeenerf} dataset.}
\vspace{-6pt}

\resizebox{\linewidth}{!}{
\begin{tabular}{l|ccccccccc}
\toprule
\begin{tabular}{c|c} Method & Scenes \end{tabular} & Amsterdam & Barcelona & Bilbao & Chicago & Hollywood & Pompidou & Quebec & Rome\\
\midrule

2D-GS\cite{huang20242d} & 1026 & 1251 & 968 & 1008 & 1125 & 1526 & 811 & 914 \\
3D-GS\cite{kerbl20233d} & 2358 & 3106 & 2190 & 2794 & 2812 & 3594 & 2176 & 2459 \\
Mip-Splatting\cite{yu2023mip} & 2325 & 2874 & 2072 & 2712 & 2578 & 3233 & 1969 & 2251 \\
Scaffold-GS\cite{lu2023scaffold} & 1219 & 1687 & 1122 & 1958 & 1117 & 2600 & 1630 & 1886 \\
\midrule
Anchor-2D-GS & 1222 & 1050 & 1054 & 1168 & 706 & 1266 & 881 & 1050 \\
Anchor-3D-GS & 1842 & 1630 & 1393 & 1593 & 1061 & 1995 & 1368 & 1641 \\
\midrule
Our-2D-GS & \textbf{703} & \textbf{771} & \textbf{629} & \textbf{631} & \textbf{680} & \textbf{786} & \textbf{582} & \textbf{629} \\
Our-3D-GS & 1094 & 1090 & 760 & 830 & 975 & 1120 & 816 & 932 \\
Our-Scaffold-GS & 1508 & 1666 & 1296 & 1284 & 1478 & 1584 & 1354 & 1622 \\
\bottomrule

\end{tabular}
}
\end{table}

\vspace{-0.5em}
\begin{table}[htbp]
\renewcommand{\arraystretch}{1.1}
\setlength{\tabcolsep}{1pt}
\centering
\caption{Storage memory(\#MB) for all scenes in the BungeeNeRF\cite{xiangli2022bungeenerf} dataset.}
\vspace{-6pt}

\resizebox{\linewidth}{!}{
\begin{tabular}{l|ccccccccc}
\toprule
\begin{tabular}{c|c} Method & Scenes \end{tabular} & Amsterdam & Barcelona & Bilbao & Chicago & Hollywood & Pompidou & Quebec & Rome\\
\midrule

2D-GS\cite{huang20242d} & 809.6 & 1027.7 & 952.2 & 633.2 & 814.3 & 1503.4 & 643.2 & 705.5 \\
3D-GS\cite{kerbl20233d} & 1569.1 & 2191.9 & 1446.1 & 1630.2 & 1758.3 & 2357.6 & 1573.7 & 1811.8 \\
Mip-Splatting\cite{yu2023mip} & 1464.3 & 1935.4 & 1341.4 & 1536.0 & 1607.7 & 2037.8 & 1382.4 & 1577.0 \\
Scaffold-GS\cite{lu2023scaffold} & 236.2 & 378.8 & 219.0 & 306.1 & 208.3 & 478.5 & 340.2 & 386.6 \\
\midrule
Anchor-2D-GS & 559.6 & 564.5 & 520.3 & 567.9 & 411.6 & 629.1 & 479.5 & 537.9 \\
Anchor-3D-GS & 866.4 & 862.8 & 699.4 & 778.5 & 607.9 & 979.3 & 725.3 & 802.5 \\
\midrule
Our-2D-GS & 449.8 & 1014.4 & 425.9 & 1127.8 & 776.2 & 765.52 & 498.8 & 830.2 \\
Our-3D-GS & 1213.5 & 1414.3 & 892.4 & 1268.5 & 960.5 & 949.8 & 618.5 & 1048.3 \\
Our-Scaffold-GS & \textbf{273.8} & \textbf{355.9} & \textbf{246.5} & \textbf{286.8} & \textbf{259.0} & \textbf{339.6} & \textbf{258.8} & \textbf{353.4}\\
\bottomrule

\end{tabular}
}
\end{table}

\vspace{-2em}
\begin{table}[htbp]
\renewcommand{\arraystretch}{1.1}
\setlength{\tabcolsep}{1pt}
\centering
\caption{PSNR for multi-resolution Mip-NeRF360\cite{barron2022mip} scenes (1$\times$ resolution).}
\vspace{-6pt}

\resizebox{\linewidth}{!}{
\begin{tabular}{l|ccccccccc}
\toprule
\begin{tabular}{c|c} Method & Scenes \end{tabular} & bicycle & bonsai & counter & flowers & garden & kitchen & room & stump & treehill\\
\midrule

3D-GS\cite{kerbl20233d} & 23.66 & 29.89 & 27.98 & 20.42 & 25.45 & 29.55 & 30.51 & 25.48 & 22.50 \\
Mip-Splatting\cite{yu2023mip} & \textbf{25.19} & 31.76 & 29.07 & \textbf{21.68} & 26.82 & 31.27 & 31.60 & \textbf{26.71} & 22.74 \\
Scaffold-GS\cite{lu2023scaffold} & 23.64 & 31.31 & 28.82 & 20.87 & 26.04 & 30.39 & 31.36 & 25.66 & 23.14 \\
\midrule
Our-Scaffold-GS & 24.21 & \textbf{33.44} & \textbf{30.15} & 20.89 & \textbf{27.01} & \textbf{31.83} & \textbf{32.39} & 25.92 & \textbf{23.26} \\
\bottomrule

\end{tabular}
}
\end{table}

\vspace{-2em}
\begin{table}[htbp]
\renewcommand{\arraystretch}{1.1}
\setlength{\tabcolsep}{1pt}
\centering
\caption{SSIM for multi-resolution Mip-NeRF360\cite{barron2022mip} scenes (1$\times$ resolution).}
\vspace{-6pt}

\resizebox{\linewidth}{!}{
\begin{tabular}{l|ccccccccc}
\toprule
\begin{tabular}{c|c} Method & Scenes \end{tabular} & bicycle & bonsai &  counter & flowers & garden & kitchen & room & stump & treehill\\
\midrule

3D-GS\cite{kerbl20233d} & 0.648 & 0.917 & 0.883 & 0.510 & 0.752 & 0.902 & 0.905 & 0.707 & 0.587 \\
Mip-Splatting\cite{yu2023mip} & \textbf{0.730} & 0.939 & 0.904 & \textbf{0.586} & 0.817 & 0.924 & 0.919 & \textbf{0.764} & 0.622 \\
Scaffold-GS\cite{lu2023scaffold} & 0.640 & 0.932 & 0.895 & 0.521 & 0.772 & 0.910 & 0.916 & 0.709 & 0.605 \\
\midrule
Our-Scaffold-GS & 0.676 & \textbf{0.952} & \textbf{0.919} & 0.541 & \textbf{0.823} & \textbf{0.930} & \textbf{0.932} & 0.722 & \textbf{0.628} \\
\bottomrule

\end{tabular}
}
\end{table}

\vspace{-2em}
\begin{table}[htbp]
\renewcommand{\arraystretch}{1.1}
\setlength{\tabcolsep}{1pt}
\centering
\caption{LPIPS for multi-resolution Mip-NeRF360\cite{barron2022mip} scenes (1$\times$ resolution).}
\vspace{-6pt}

\resizebox{\linewidth}{!}{
\begin{tabular}{l|ccccccccc}
\toprule
\begin{tabular}{c|c} Method & Scenes \end{tabular} & bicycle & bonsai &  counter & flowers & garden & kitchen & room & stump & treehill\\
\midrule

3D-GS\cite{kerbl20233d} & 0.359 & 0.223 & 0.235 & 0.443 & 0.269 & 0.167 & 0.242 & 0.331 & 0.440 \\
Mip-Splatting\cite{yu2023mip} & \textbf{0.275} & 0.188 & 0.196 & \textbf{0.367} & 0.190 & 0.130 & 0.214 & \textbf{0.258} & 0.379 \\
Scaffold-GS\cite{lu2023scaffold} & 0.355 & 0.208 & 0.219 & 0.430 & 0.242 & 0.159 & 0.219 & 0.326 & 0.407 \\
\midrule
Our-Scaffold-GS & 0.313 & \textbf{0.169} & \textbf{0.178} & 0.401 & \textbf{0.168} & \textbf{0.119} & \textbf{0.186} & 0.309 & \textbf{0.364} \\
\bottomrule

\end{tabular}
}
\end{table}

\vspace{-2em}
\begin{table}[htbp]
\renewcommand{\arraystretch}{1.1}
\setlength{\tabcolsep}{1pt}
\centering
\caption{PSNR for multi-resolution Mip-NeRF360\cite{barron2022mip} scenes (2$\times$ resolution).}
\vspace{-6pt}

\resizebox{\linewidth}{!}{
\begin{tabular}{l|ccccccccc}
\toprule
\begin{tabular}{c|c} Method & Scenes \end{tabular} & bicycle & garden & stump & room & counter & kitchen & bonsai & flowers & treehill\\
\midrule

3D-GS\cite{kerbl20233d} & 25.41 & 27.56 & 26.42 & 31.29 & 28.57 & 30.54 & 30.71 & 21.83 & 23.67 \\
Mip-Splatting\cite{yu2023mip} & \textbf{26.83} & 28.80 & 27.57 & 32.44 & 29.59 & 32.27 & 32.41 & \textbf{23.22} & 23.90 \\
Scaffold-GS\cite{lu2023scaffold} & 25.43 & 28.37 & 26.60 & 32.36 & 29.52 & 31.50 & 32.20 & 22.36 & 24.51 \\
\midrule
Our-Scaffold-GS & 25.92 & \textbf{29.08} & 26.81 & \textbf{33.31} & \textbf{30.77} & \textbf{32.44} & \textbf{34.13} & 22.38 & \textbf{24.53} \\
\bottomrule

\end{tabular}
}
\end{table}

\vspace{-2em}
\begin{table}[htbp]
\renewcommand{\arraystretch}{1.1}
\setlength{\tabcolsep}{1pt}
\centering
\caption{SSIM for multi-resolution Mip-NeRF360\cite{barron2022mip} scenes (2$\times$ resolution).}
\vspace{-6pt}

\resizebox{\linewidth}{!}{
\begin{tabular}{l|ccccccccc}
\toprule
\begin{tabular}{c|c} Method & Scenes \end{tabular} & bicycle & garden & stump & room & counter & kitchen & bonsai & flowers & treehill\\
\midrule

3D-GS\cite{kerbl20233d} & 0.756 & 0.866 & 0.769 & 0.933 & 0.904 & 0.935 & 0.939 & 0.620 & 0.676 \\
Mip-Splatting\cite{yu2023mip} & \textbf{0.823} & 0.902 & \textbf{0.819} & 0.946 & 0.923 & \textbf{0.950} & \textbf{0.956} & \textbf{0.693} & 0.705 \\
Scaffold-GS\cite{lu2023scaffold} & 0.759 & 0.883 & 0.773 & 0.946 & 0.918 & 0.941 & 0.953 & 0.640 & 0.701 \\
\midrule
Our-Scaffold-GS & 0.785 & \textbf{0.903} & 0.781 & \textbf{0.956} & \textbf{0.937} & 0.949 & \textbf{0.966} & 0.657 & \textbf{0.714} \\
\bottomrule

\end{tabular}
}
\end{table}

\vspace{-2em}
\begin{table}[htbp]
\renewcommand{\arraystretch}{1.1}
\setlength{\tabcolsep}{1pt}
\centering
\caption{LPIPS for multi-resolution Mip-NeRF360\cite{barron2022mip} scenes (2$\times$ resolution).}
\vspace{-6pt}

\resizebox{\linewidth}{!}{
\begin{tabular}{l|ccccccccc}
\toprule
\begin{tabular}{c|c} Method & Scenes \end{tabular} & bicycle & garden & stump & room & counter & kitchen & bonsai & flowers & treehill\\
\midrule

3D-GS\cite{kerbl20233d} & 0.261 & 0.138 & 0.239 & 0.134 & 0.141 & 0.093 & 0.114 & 0.351 & 0.349\\
Mip-Splatting\cite{yu2023mip} & \textbf{0.177} & 0.084 & \textbf{0.170} & 0.110 & 0.110 & \textbf{0.067} & 0.088 & \textbf{0.276} & 0.284\\
Scaffold-GS\cite{lu2023scaffold} & 0.245 & 0.110 & 0.234 & 0.108 & 0.125 & 0.086 & 0.099 & 0.335 & 0.307\\
\midrule
Our-Scaffold-GS & 0.210 & \textbf{0.080} & 0.221 & \textbf{0.087} & \textbf{0.095} & 0.068 & \textbf{0.071} & 0.304 & \textbf{0.274}\\
\bottomrule

\end{tabular}
}
\end{table}

\vspace{-2em}
\begin{table}[htbp]
\renewcommand{\arraystretch}{1.1}
\setlength{\tabcolsep}{1pt}
\centering
\caption{PSNR for multi-resolution Mip-NeRF360\cite{barron2022mip} scenes (4$\times$ resolution).}
\vspace{-6pt}

\resizebox{\linewidth}{!}{
\begin{tabular}{l|ccccccccc}
\toprule
\begin{tabular}{c|c} Method & Scenes \end{tabular} & bicycle & garden & stump & room & counter & kitchen & bonsai & flowers & treehill\\
\midrule

3D-GS\cite{kerbl20233d} & 27.06 & 29.19 & 27.77 & 31.75 & 29.29 & 31.51 & 31.25 & 24.04 & 25.12 \\
Mip-Splatting\cite{yu2023mip} & 28.66 & 30.69 & 29.12 & 33.29 & 30.44 & \textbf{33.40} & 33.25 & 25.66 & 25.53 \\
Scaffold-GS\cite{lu2023scaffold} & 27.34 & 30.40 & 28.11 & 33.03 & 30.42 & 32.55 & 32.83 & 24.72 & \textbf{26.31} \\
\midrule
Our-Scaffold-GS & 28.00 & \textbf{31.23} & 28.36 & \textbf{34.01} & \textbf{31.60} & 33.39 & \textbf{34.86} & 24.66 & 26.27 \\
\bottomrule

\end{tabular}
}
\end{table}

\vspace{-2em}
\begin{table}[htbp]
\renewcommand{\arraystretch}{1.1}
\setlength{\tabcolsep}{1pt}
\centering
\caption{SSIM for multi-resolution Mip-NeRF360\cite{barron2022mip} scenes (4$\times$ resolution).}
\vspace{-6pt}

\resizebox{\linewidth}{!}{
\begin{tabular}{l|ccccccccc}
\toprule
\begin{tabular}{c|c} Method & Scenes \end{tabular} & bicycle & garden & stump & room & counter & kitchen & bonsai & flowers & treehill\\
\midrule

3D-GS\cite{kerbl20233d} & 0.857 & 0.921 & 0.841 & 0.954 & 0.929 & 0.958 & 0.953 & 0.753 & 0.788 \\
Mip-Splatting\cite{yu2023mip} & \textbf{0.901} & \textbf{0.945} & \textbf{0.882} & 0.965 & 0.943 & \textbf{0.967} & 0.968 & \textbf{0.807} & 0.811 \\
Scaffold-GS\cite{lu2023scaffold} & 0.868 & 0.936 & 0.852 & 0.966 & 0.942 & 0.963 & 0.966 & 0.776 & \textbf{0.815} \\
\midrule
Our-Scaffold-GS & 0.883 & \textbf{0.945} & 0.857 & \textbf{0.971} & \textbf{0.952} & 0.966 & \textbf{0.975} & 0.782 & 0.822 \\
\bottomrule

\end{tabular}
}
\end{table}

\vspace{-2em}
\begin{table}[htbp]
\renewcommand{\arraystretch}{1.1}
\setlength{\tabcolsep}{1pt}
\centering
\caption{LPIPS for multi-resolution Mip-NeRF360\cite{barron2022mip} scenes (4$\times$ resolution).}
\vspace{-6pt}

\resizebox{\linewidth}{!}{
\begin{tabular}{l|ccccccccc}
\toprule
\begin{tabular}{c|c} Method & Scenes \end{tabular} & bicycle & garden & stump & room & counter & kitchen & bonsai & flowers & treehill\\
\midrule

3D-GS\cite{kerbl20233d} & 0.140 & 0.062 & 0.149 & 0.066 & 0.081 & 0.045 & 0.059 & 0.227 & 0.220 \\
Mip-Splatting\cite{yu2023mip} & \textbf{0.085} & 0.040 & \textbf{0.102} & 0.050 & 0.063 & 0.038 & \textbf{0.043} & \textbf{0.177} & 0.183 \\
Scaffold-GS\cite{lu2023scaffold} & 0.118 & 0.048 & 0.138 & 0.047 & 0.069 & 0.039 & 0.045 & 0.204 & 0.185 \\
\midrule
Our-Scaffold-GS & 0.101 & \textbf{0.039} & 0.131 & \textbf{0.039} & \textbf{0.054} & \textbf{0.036} & \textbf{0.032} & 0.182 & \textbf{0.168} \\
\bottomrule

\end{tabular}
}
\end{table}

\vspace{-2em}
\begin{table}[htbp]
\renewcommand{\arraystretch}{1.1}
\setlength{\tabcolsep}{1pt}
\centering
\caption{PSNR for multi-resolution Mip-NeRF360\cite{barron2022mip} scenes (8$\times$ resolution).}
\vspace{-6pt}

\resizebox{\linewidth}{!}{
\begin{tabular}{l|ccccccccc}
\toprule
\begin{tabular}{c|c} Method & Scenes \end{tabular} & bicycle & garden & stump & room & counter & kitchen & bonsai & flowers & treehill\\
\midrule

3D-GS\cite{kerbl20233d} & 26.26 & 29.28 & 27.50 & 30.45 & 28.14 & 29.86 & 29.25 & 24.33 & 25.62\\
Mip-Splatting\cite{yu2023mip} & \textbf{29.80} & 31.93 & \textbf{30.78} & 33.60 & 31.11 & \textbf{33.74} & 33.38 & \textbf{27.95} & 27.13\\
Scaffold-GS\cite{lu2023scaffold} & 27.29 & 30.26 & 28.61 & 31.51 & 29.67 & 30.84 & 30.61 & 24.99 & 27.04\\
\midrule
Our-Scaffold-GS & 29.09 & \textbf{32.61} & 29.05 & \textbf{34.24} & \textbf{32.35} & 34.35 & \textbf{35.42} & 25.83 & \textbf{27.69}\\
\bottomrule

\end{tabular}
}
\end{table}

\vspace{-2em}
\begin{table}[htbp]
\renewcommand{\arraystretch}{1.1}
\setlength{\tabcolsep}{1pt}
\centering
\caption{SSIM for multi-resolution Mip-NeRF360\cite{barron2022mip} scenes (8$\times$ resolution).}
\vspace{-6pt}

\resizebox{\linewidth}{!}{
\begin{tabular}{l|ccccccccc}
\toprule
\begin{tabular}{c|c} Method & Scenes \end{tabular} & bicycle & garden & stump & room & counter & kitchen & bonsai & flowers & treehill\\
\midrule

3D-GS\cite{kerbl20233d} & 0.871 & 0.930 & 0.846 & 0.953 & 0.928 & 0.954 & 0.944 & 0.805 & 0.840\\
Mip-Splatting\cite{yu2023mip} & \textbf{0.938} & \textbf{0.964} & \textbf{0.925} & 0.973 & 0.957 & 0.975 & 0.973 & \textbf{0.883} & \textbf{0.886}\\
Scaffold-GS\cite{lu2023scaffold} & 0.894 & 0.941 & 0.875 & 0.965 & 0.946 & 0.961 & 0.959 & 0.825 & 0.871\\
\midrule
Our-Scaffold-GS & 0.919 & \textbf{0.964} & 0.885 & \textbf{0.978} & \textbf{0.964} & \textbf{0.977} & \textbf{0.981} & 0.838 & 0.885\\
\bottomrule

\end{tabular}
}
\end{table}

\vspace{-2em}
\begin{table}[htbp]
\renewcommand{\arraystretch}{1.1}
\setlength{\tabcolsep}{1pt}
\centering
\caption{LPIPS for multi-resolution Mip-NeRF360\cite{barron2022mip} scenes (8$\times$ resolution).}
\vspace{-6pt}

\resizebox{\linewidth}{!}{
\begin{tabular}{l|ccccccccc}
\toprule
\begin{tabular}{c|c} Method & Scenes \end{tabular} & bicycle & garden & stump & room & counter & kitchen & bonsai & flowers & treehill\\
\midrule

3D-GS\cite{kerbl20233d} & 0.098 & 0.047 & 0.126 & 0.048 & 0.063 & 0.037 & 0.047 & 0.159 & 0.147\\
Mip-Splatting\cite{yu2023mip} & \textbf{0.049} & \textbf{0.026} & \textbf{0.068} & 0.031 & 0.041 & 0.029 & 0.029 & \textbf{0.109} & 0.113\\
Scaffold-GS\cite{lu2023scaffold} & 0.082 & 0.040 & 0.110 & 0.033 & 0.048 & 0.032 & 0.035 & 0.144 & 0.120\\
\midrule
Our-Scaffold-GS & 0.062 & \textbf{0.025} & 0.103 & \textbf{0.023} & \textbf{0.032} & \textbf{0.021} & \textbf{0.017} & 0.118 & \textbf{0.106}\\
\bottomrule

\end{tabular}
}
\end{table}

\vspace{-2em}
\begin{table}[htbp]
\centering
\caption{Quantitative results for multi-resolution Tanks\&Temples\cite{knapitsch2017tanks} dataset.}
\vspace{-6pt}
\renewcommand{\arraystretch}{1.1}
\setlength{\tabcolsep}{4pt}
\resizebox{\linewidth}{!}{
\begin{tabular}{l|cccc|cccc}
\toprule
{\textbf{PSNR}} & \multicolumn{4}{c|}{Train}  & \multicolumn{4}{c}{Truck}\\ 
\begin{tabular}{c|c} Method & Scales \end{tabular} & $1 \times$ & $2 \times$ & $4 \times$ & $8 \times$ & $1 \times$ & $2 \times$ & $4 \times$ & $8 \times$  \\
\midrule

3D-GS\cite{kerbl20233d} & 21.23 & 22.17 & 22.69 & 22.16 & 23.92 & 25.47 & 26.24 & 25.51 \\
Mip-Splatting\cite{yu2023mip} & 21.87 & 22.70 & 23.41 & 23.83 & 25.29 & 26.79 & 28.07 & 28.81 \\
Scaffold-GS\cite{lu2023scaffold} & 21.91 & 23.04 & 23.84 & 23.50 & 24.66 & 26.47 & 27.44 & 26.67 \\
\midrule
Our-Scaffold-GS & \textbf{22.49} & \textbf{23.50} & \textbf{24.18} & \textbf{24.22} & \textbf{25.85} & \textbf{27.53} & \textbf{28.83} & \textbf{29.67} \\
\bottomrule

\end{tabular}
}

\vspace{1em}
\resizebox{\linewidth}{!}{
\begin{tabular}{l|cccc|cccc}
\toprule
{\textbf{SSIM}} & \multicolumn{4}{c|}{Train}  & \multicolumn{4}{c}{Truck}\\ 
\begin{tabular}{c|c} Method & Scales \end{tabular} & $1 \times$ & $2 \times$ & $4 \times$ & $8 \times$ & $1 \times$ & $2 \times$ & $4 \times$ & $8 \times$  \\
\midrule

3D-GS\cite{kerbl20233d} & 0.754 & 0.830 & 0.879 & 0.880 & 0.827 & 0.899 & 0.930 & 0.929 \\
Mip-Splatting\cite{yu2023mip} & 0.791 & 0.859 & 0.906 & 0.929 & 0.868 & 0.925 & 0.955 & 0.969 \\
Scaffold-GS\cite{lu2023scaffold} & 0.781 & 0.860 & 0.907 & 0.913 & 0.844 & 0.916 & 0.946 & 0.945 \\
\midrule
Our-Scaffold-GS & \textbf{0.817} & \textbf{0.882} & \textbf{0.919} & \textbf{0.932} & \textbf{0.878} & \textbf{0.932} & \textbf{0.958} & \textbf{0.971} \\
\bottomrule

\end{tabular}
}

\vspace{1em}
\resizebox{\linewidth}{!}{
\begin{tabular}{l|cccc|cccc}
\toprule
{\textbf{LPIPS}} & \multicolumn{4}{c|}{Train}  & \multicolumn{4}{c}{Truck}\\ 
\begin{tabular}{c|c} Method & Scales \end{tabular} & $1 \times$ & $2 \times$ & $4 \times$ & $8 \times$ & $1 \times$ & $2 \times$ & $4 \times$ & $8 \times$  \\
\midrule

3D-GS\cite{kerbl20233d} & 0.292 & 0.181 & 0.106 & 0.093 & 0.239 & 0.116 & 0.058 & 0.050 \\
Mip-Splatting\cite{yu2023mip} & 0.243 & 0.143 & 0.080 & 0.056 & 0.179 & 0.082 & 0.039 & 0.025 \\
Scaffold-GS\cite{lu2023scaffold} & 0.261 & 0.149 & 0.080 & 0.070 & 0.216 & 0.094 & 0.045 & 0.041 \\
\midrule
Our-Scaffold-GS & \textbf{0.216} & \textbf{0.119} & \textbf{0.068} & \textbf{0.055} & \textbf{0.154} & \textbf{0.066} & \textbf{0.033} & \textbf{0.023} \\
\bottomrule

\end{tabular}
}
\end{table}

\vspace{-2em}
\begin{table}[htbp]
\centering
\caption{Quantitative results for multi-resolution Deep Blending\cite{hedman2018deep} dataset.}
\vspace{-6pt}
\renewcommand{\arraystretch}{1.1}
\setlength{\tabcolsep}{4pt}
\resizebox{\linewidth}{!}{
\begin{tabular}{l|cccc|cccc}
\toprule
{\textbf{PSNR}} & \multicolumn{4}{c|}{Dr Johnson}  & \multicolumn{4}{c}{Playroom}\\ 
\begin{tabular}{c|c} Method & Scales \end{tabular} & $1 \times$ & $2 \times$ & $4 \times$ & $8 \times$ & $1 \times$ & $2 \times$ & $4 \times$ & $8 \times$  \\
\midrule

3D-GS\cite{kerbl20233d} & 28.62 & 28.97 & 29.23 & 28.71 & 29.43 & 29.89 & 30.25 & 29.47 \\
Mip-Splatting\cite{yu2023mip} & 28.95 & 29.30 & 29.91 & 30.55 & 30.18 & 30.62 & 31.16 & 31.61 \\
Scaffold-GS\cite{lu2023scaffold} & 29.51 & 29.99 & \textbf{30.58} & 30.31 & 29.77 & 30.39 & 31.10 & 30.47 \\
\midrule
Our-Scaffold-GS & \textbf{29.75} & \textbf{30.14} & \textbf{30.58} & \textbf{30.92} & \textbf{30.87} & \textbf{31.42} & \textbf{31.76} & \textbf{31.63} \\
\bottomrule

\end{tabular}
}

\vspace{1em}
\resizebox{\linewidth}{!}{
\begin{tabular}{l|cccc|cccc}
\toprule
{\textbf{SSIM}} & \multicolumn{4}{c|}{Dr Johnson}  & \multicolumn{4}{c}{Playroom}\\ 
\begin{tabular}{c|c} Method & Scales \end{tabular} & $1 \times$ & $2 \times$ & $4 \times$ & $8 \times$ & $1 \times$ & $2 \times$ & $4 \times$ & $8 \times$  \\
\midrule

3D-GS\cite{kerbl20233d} & 0.890 & 0.900 & 0.911 & 0.907 & 0.898 & 0.919 & 0.935 & 0.934 \\
Mip-Splatting\cite{yu2023mip} & 0.900 & 0.911 & 0.925 & 0.936 & 0.909 & 0.929 & 0.946 & 0.956 \\
Scaffold-GS\cite{lu2023scaffold} & 0.900 & 0.914 & 0.930 & 0.932 & 0.900 & 0.923 & 0.944 & 0.949 \\
\midrule
Our-Scaffold-GS & \textbf{0.908} & \textbf{0.920} & \textbf{0.932} & \textbf{0.940} & \textbf{0.911} & \textbf{0.933} & \textbf{0.949} & \textbf{0.957} \\
\bottomrule

\end{tabular}
}

\vspace{1em}
\resizebox{\linewidth}{!}{
\begin{tabular}{l|cccc|cccc}
\toprule
{\textbf{LPIPS}} & \multicolumn{4}{c|}{Dr Johnson}  & \multicolumn{4}{c}{Playroom}\\ 
\begin{tabular}{c|c} Method & Scales \end{tabular} & $1 \times$ & $2 \times$ & $4 \times$ & $8 \times$ & $1 \times$ & $2 \times$ & $4 \times$ & $8 \times$  \\
\midrule

3D-GS\cite{kerbl20233d} & 0.277 & 0.177 & 0.103 & 0.083 & 0.277 & 0.170 & 0.081 & 0.060 \\
Mip-Splatting\cite{yu2023mip} & 0.251 & 0.151 & 0.084 & 0.060 & \textbf{0.247} & \textbf{0.140} & \textbf{0.061} & 0.039 \\
Scaffold-GS\cite{lu2023scaffold} & \textbf{0.244} & \textbf{0.144} & \textbf{0.078} & \textbf{0.057} & 0.257 & 0.150 & 0.064 & \textbf{0.038} \\
\midrule
Our-Scaffold-GS & 0.263 & 0.159 & 0.082 & 0.061 & 0.274 & 0.164 & 0.068 & 0.041 \\
\bottomrule

\end{tabular}
}
\end{table}

\newpage
\bibliographystyle{IEEEtran}
\bibliography{reference} 

\begin{thebibliography}{10}
\providecommand{\url}[1]{#1}
\csname url@samestyle\endcsname
\providecommand{\newblock}{\relax}
\providecommand{\bibinfo}[2]{#2}
\providecommand{\BIBentrySTDinterwordspacing}{\spaceskip=0pt\relax}
\providecommand{\BIBentryALTinterwordstretchfactor}{4}
\providecommand{\BIBentryALTinterwordspacing}{\spaceskip=\fontdimen2\font plus
\BIBentryALTinterwordstretchfactor\fontdimen3\font minus \fontdimen4\font\relax}
\providecommand{\BIBforeignlanguage}[2]{{%
\expandafter\ifx\csname l@#1\endcsname\relax
\typeout{** WARNING: IEEEtran.bst: No hyphenation pattern has been}%
\typeout{** loaded for the language `#1'. Using the pattern for}%
\typeout{** the default language instead.}%
\else
\language=\csname l@#1\endcsname
\fi
#2}}
\providecommand{\BIBdecl}{\relax}
\BIBdecl

\bibitem{li2023matrixcity}
Y.~Li, L.~Jiang, L.~Xu, Y.~Xiangli, Z.~Wang, D.~Lin, and B.~Dai, ``Matrixcity: A large-scale city dataset for city-scale neural rendering and beyond,'' in \emph{Proceedings of the IEEE/CVF International Conference on Computer Vision}, 2023, pp. 3205--3215.

\bibitem{kerbl2024hierarchical}
B.~Kerbl, A.~Meuleman, G.~Kopanas, M.~Wimmer, A.~Lanvin, and G.~Drettakis, ``A hierarchical 3d gaussian representation for real-time rendering of very large datasets,'' \emph{ACM Transactions on Graphics (TOG)}, vol.~43, no.~4, pp. 1--15, 2024.

\bibitem{lu2023scaffold}
T.~Lu, M.~Yu, L.~Xu, Y.~Xiangli, L.~Wang, D.~Lin, and B.~Dai, ``Scaffold-gs: Structured 3d gaussians for view-adaptive rendering,'' \emph{arXiv preprint arXiv:2312.00109}, 2023.

\bibitem{mildenhall2021nerf}
B.~Mildenhall, P.~P. Srinivasan, M.~Tancik, J.~T. Barron, R.~Ramamoorthi, and R.~Ng, ``Nerf: Representing scenes as neural radiance fields for view synthesis,'' \emph{Communications of the ACM}, vol.~65, no.~1, pp. 99--106, 2021.

\bibitem{kerbl20233d}
B.~Kerbl, G.~Kopanas, T.~Leimk{\"u}hler, and G.~Drettakis, ``3d gaussian splatting for real-time radiance field rendering,'' \emph{ACM Transactions on Graphics}, vol.~42, no.~4, 2023.

\bibitem{zielonka2023drivable}
W.~Zielonka, T.~Bagautdinov, S.~Saito, M.~Zollh{\"o}fer, J.~Thies, and J.~Romero, ``Drivable 3d gaussian avatars,'' \emph{arXiv preprint arXiv:2311.08581}, 2023.

\bibitem{saito2023relightable}
S.~Saito, G.~Schwartz, T.~Simon, J.~Li, and G.~Nam, ``Relightable gaussian codec avatars,'' \emph{arXiv preprint arXiv:2312.03704}, 2023.

\bibitem{zheng2023gps}
S.~Zheng, B.~Zhou, R.~Shao, B.~Liu, S.~Zhang, L.~Nie, and Y.~Liu, ``Gps-gaussian: Generalizable pixel-wise 3d gaussian splatting for real-time human novel view synthesis,'' \emph{arXiv preprint arXiv:2312.02155}, 2023.

\bibitem{qian2023gaussianavatars}
S.~Qian, T.~Kirschstein, L.~Schoneveld, D.~Davoli, S.~Giebenhain, and M.~Nie{\ss}ner, ``Gaussianavatars: Photorealistic head avatars with rigged 3d gaussians,'' \emph{arXiv preprint arXiv:2312.02069}, 2023.

\bibitem{yan2024street}
Y.~Yan, H.~Lin, C.~Zhou, W.~Wang, H.~Sun, K.~Zhan, X.~Lang, X.~Zhou, and S.~Peng, ``Street gaussians for modeling dynamic urban scenes,'' \emph{arXiv preprint arXiv:2401.01339}, 2024.

\bibitem{zhou2023drivinggaussian}
X.~Zhou, Z.~Lin, X.~Shan, Y.~Wang, D.~Sun, and M.-H. Yang, ``Drivinggaussian: Composite gaussian splatting for surrounding dynamic autonomous driving scenes,'' \emph{arXiv preprint arXiv:2312.07920}, 2023.

\bibitem{jiang2024vr}
Y.~Jiang, C.~Yu, T.~Xie, X.~Li, Y.~Feng, H.~Wang, M.~Li, H.~Lau, F.~Gao, Y.~Yang \emph{et~al.}, ``Vr-gs: A physical dynamics-aware interactive gaussian splatting system in virtual reality,'' \emph{arXiv preprint arXiv:2401.16663}, 2024.

\bibitem{xie2023physgaussian}
T.~Xie, Z.~Zong, Y.~Qiu, X.~Li, Y.~Feng, Y.~Yang, and C.~Jiang, ``Physgaussian: Physics-integrated 3d gaussians for generative dynamics,'' \emph{arXiv preprint arXiv:2311.12198}, 2023.

\bibitem{yu2023mip}
Z.~Yu, A.~Chen, B.~Huang, T.~Sattler, and A.~Geiger, ``Mip-splatting: Alias-free 3d gaussian splatting,'' \emph{arXiv preprint arXiv:2311.16493}, 2023.

\bibitem{huang20242d}
B.~Huang, Z.~Yu, A.~Chen, A.~Geiger, and S.~Gao, ``2d gaussian splatting for geometrically accurate radiance fields,'' in \emph{ACM SIGGRAPH 2024 Conference Papers}, 2024, pp. 1--11.

\bibitem{xu2023vr}
L.~Xu, V.~Agrawal, W.~Laney, T.~Garcia, A.~Bansal, C.~Kim, S.~Rota~Bul{\`o}, L.~Porzi, P.~Kontschieder, A.~Bo{\v{z}}i{\v{c}} \emph{et~al.}, ``Vr-nerf: High-fidelity virtualized walkable spaces,'' in \emph{SIGGRAPH Asia 2023 Conference Papers}, 2023, pp. 1--12.

\bibitem{yu2021plenoctrees}
A.~Yu, R.~Li, M.~Tancik, H.~Li, R.~Ng, and A.~Kanazawa, ``Plenoctrees for real-time rendering of neural radiance fields,'' in \emph{Proceedings of the IEEE/CVF International Conference on Computer Vision}, 2021, pp. 5752--5761.

\bibitem{martel2021acorn}
J.~N. Martel, D.~B. Lindell, C.~Z. Lin, E.~R. Chan, M.~Monteiro, and G.~Wetzstein, ``Acorn: Adaptive coordinate networks for neural scene representation,'' \emph{arXiv preprint arXiv:2105.02788}, 2021.

\bibitem{liu2024citygaussian}
Y.~Liu, H.~Guan, C.~Luo, L.~Fan, J.~Peng, and Z.~Zhang, ``Citygaussian: Real-time high-quality large-scale scene rendering with gaussians,'' \emph{arXiv preprint arXiv:2404.01133}, 2024.

\bibitem{liu2020neural}
L.~Liu, J.~Gu, K.~Zaw~Lin, T.-S. Chua, and C.~Theobalt, ``Neural sparse voxel fields,'' \emph{Advances in Neural Information Processing Systems}, vol.~33, pp. 15\,651--15\,663, 2020.

\bibitem{fridovich2022plenoxels}
S.~Fridovich-Keil, A.~Yu, M.~Tancik, Q.~Chen, B.~Recht, and A.~Kanazawa, ``Plenoxels: Radiance fields without neural networks,'' in \emph{Proceedings of the IEEE/CVF Conference on Computer Vision and Pattern Recognition}, 2022, pp. 5501--5510.

\bibitem{sun2022direct}
C.~Sun, M.~Sun, and H.-T. Chen, ``Direct voxel grid optimization: Super-fast convergence for radiance fields reconstruction,'' in \emph{Proceedings of the IEEE/CVF Conference on Computer Vision and Pattern Recognition}, 2022, pp. 5459--5469.

\bibitem{chen2022tensorf}
A.~Chen, Z.~Xu, A.~Geiger, J.~Yu, and H.~Su, ``Tensorf: Tensorial radiance fields,'' in \emph{European Conference on Computer Vision}.\hskip 1em plus 0.5em minus 0.4em\relax Springer, 2022, pp. 333--350.

\bibitem{muller2022instant}
T.~M{\"u}ller, A.~Evans, C.~Schied, and A.~Keller, ``Instant neural graphics primitives with a multiresolution hash encoding,'' \emph{ACM Transactions on Graphics (ToG)}, vol.~41, no.~4, pp. 1--15, 2022.

\bibitem{xu2023grid}
L.~Xu, Y.~Xiangli, S.~Peng, X.~Pan, N.~Zhao, C.~Theobalt, B.~Dai, and D.~Lin, ``Grid-guided neural radiance fields for large urban scenes,'' in \emph{Proceedings of the IEEE/CVF Conference on Computer Vision and Pattern Recognition}, 2023, pp. 8296--8306.

\bibitem{xiangli2023assetfield}
Y.~Xiangli, L.~Xu, X.~Pan, N.~Zhao, B.~Dai, and D.~Lin, ``Assetfield: Assets mining and reconfiguration in ground feature plane representation,'' \emph{arXiv preprint arXiv:2303.13953}, 2023.

\bibitem{turki2024pynerf}
H.~Turki, M.~Zollh{\"o}fer, C.~Richardt, and D.~Ramanan, ``Pynerf: Pyramidal neural radiance fields,'' \emph{Advances in Neural Information Processing Systems}, vol.~36, 2024.

\bibitem{li2023neuralangelo}
Z.~Li, T.~M{\"u}ller, A.~Evans, R.~H. Taylor, M.~Unberath, M.-Y. Liu, and C.-H. Lin, ``Neuralangelo: High-fidelity neural surface reconstruction,'' in \emph{Proceedings of the IEEE/CVF Conference on Computer Vision and Pattern Recognition}, 2023, pp. 8456--8465.

\bibitem{reiser2024binary}
C.~Reiser, S.~Garbin, P.~P. Srinivasan, D.~Verbin, R.~Szeliski, B.~Mildenhall, J.~T. Barron, P.~Hedman, and A.~Geiger, ``Binary opacity grids: Capturing fine geometric detail for mesh-based view synthesis,'' \emph{arXiv preprint arXiv:2402.12377}, 2024.

\bibitem{barron2023zip}
J.~T. Barron, B.~Mildenhall, D.~Verbin, P.~P. Srinivasan, and P.~Hedman, ``Zip-nerf: Anti-aliased grid-based neural radiance fields,'' \emph{arXiv preprint arXiv:2304.06706}, 2023.

\bibitem{tang2023dreamgaussian}
J.~Tang, J.~Ren, H.~Zhou, Z.~Liu, and G.~Zeng, ``Dreamgaussian: Generative gaussian splatting for efficient 3d content creation,'' \emph{arXiv preprint arXiv:2309.16653}, 2023.

\bibitem{liang2023luciddreamer}
Y.~Liang, X.~Yang, J.~Lin, H.~Li, X.~Xu, and Y.~Chen, ``Luciddreamer: Towards high-fidelity text-to-3d generation via interval score matching,'' \emph{arXiv preprint arXiv:2311.11284}, 2023.

\bibitem{tang2024lgm}
J.~Tang, Z.~Chen, X.~Chen, T.~Wang, G.~Zeng, and Z.~Liu, ``Lgm: Large multi-view gaussian model for high-resolution 3d content creation,'' \emph{arXiv preprint arXiv:2402.05054}, 2024.

\bibitem{feng2024gaussian}
Y.~Feng, X.~Feng, Y.~Shang, Y.~Jiang, C.~Yu, Z.~Zong, T.~Shao, H.~Wu, K.~Zhou, C.~Jiang \emph{et~al.}, ``Gaussian splashing: Dynamic fluid synthesis with gaussian splatting,'' \emph{arXiv preprint arXiv:2401.15318}, 2024.

\bibitem{luiten2023dynamic}
J.~Luiten, G.~Kopanas, B.~Leibe, and D.~Ramanan, ``Dynamic 3d gaussians: Tracking by persistent dynamic view synthesis,'' \emph{arXiv preprint arXiv:2308.09713}, 2023.

\bibitem{yang2023deformable}
Z.~Yang, X.~Gao, W.~Zhou, S.~Jiao, Y.~Zhang, and X.~Jin, ``Deformable 3d gaussians for high-fidelity monocular dynamic scene reconstruction,'' \emph{arXiv preprint arXiv:2309.13101}, 2023.

\bibitem{huang2023sc}
Y.-H. Huang, Y.-T. Sun, Z.~Yang, X.~Lyu, Y.-P. Cao, and X.~Qi, ``Sc-gs: Sparse-controlled gaussian splatting for editable dynamic scenes,'' \emph{arXiv preprint arXiv:2312.14937}, 2023.

\bibitem{yugay2023gaussian}
V.~Yugay, Y.~Li, T.~Gevers, and M.~R. Oswald, ``Gaussian-slam: Photo-realistic dense slam with gaussian splatting,'' \emph{arXiv preprint arXiv:2312.10070}, 2023.

\bibitem{keetha2023splatam}
N.~Keetha, J.~Karhade, K.~M. Jatavallabhula, G.~Yang, S.~Scherer, D.~Ramanan, and J.~Luiten, ``Splatam: Splat, track \& map 3d gaussians for dense rgb-d slam,'' \emph{arXiv preprint arXiv:2312.02126}, 2023.

\bibitem{xu2022point}
Q.~Xu, Z.~Xu, J.~Philip, S.~Bi, Z.~Shu, K.~Sunkavalli, and U.~Neumann, ``Point-nerf: Point-based neural radiance fields,'' in \emph{Proceedings of the IEEE/CVF Conference on Computer Vision and Pattern Recognition}, 2022, pp. 5438--5448.

\bibitem{fridovich2023k}
S.~Fridovich-Keil, G.~Meanti, F.~R. Warburg, B.~Recht, and A.~Kanazawa, ``K-planes: Explicit radiance fields in space, time, and appearance,'' in \emph{Proceedings of the IEEE/CVF Conference on Computer Vision and Pattern Recognition}, 2023, pp. 12\,479--12\,488.

\bibitem{cao2023hexplane}
A.~Cao and J.~Johnson, ``Hexplane: A fast representation for dynamic scenes,'' in \emph{Proceedings of the IEEE/CVF Conference on Computer Vision and Pattern Recognition}, 2023, pp. 130--141.

\bibitem{rubin19803}
S.~M. Rubin and T.~Whitted, ``A 3-dimensional representation for fast rendering of complex scenes,'' in \emph{Proceedings of the 7th annual conference on Computer graphics and interactive techniques}, 1980, pp. 110--116.

\bibitem{laine2010efficient}
S.~Laine and T.~Karras, ``Efficient sparse voxel octrees--analysis, extensions, and implementation,'' \emph{NVIDIA Corporation}, vol.~2, no.~6, 2010.

\bibitem{bai2023dynamic}
H.~Bai, Y.~Lin, Y.~Chen, and L.~Wang, ``Dynamic plenoctree for adaptive sampling refinement in explicit nerf,'' in \emph{Proceedings of the IEEE/CVF International Conference on Computer Vision}, 2023, pp. 8785--8795.

\bibitem{verdie_tog15}
Y.~Verdie, F.~Lafarge, and P.~Alliez, ``{LOD Generation for Urban Scenes},'' \emph{ACM Trans. on Graphics}, vol.~34, no.~3, 2015.

\bibitem{fang_cvpr18}
H.~Fang, F.~Lafarge, and M.~Desbrun, ``{Planar Shape Detection at Structural Scales},'' in \emph{Proc. of the IEEE conference on Computer Vision and Pattern Recognition (CVPR)}, Salt Lake City, US, 2018.

\bibitem{Yu_cvpr22}
M.~Yu and F.~Lafarge, ``{Finding Good Configurations of Planar Primitives in Unorganized Point Clouds},'' in \emph{Proc. of the IEEE conference on Computer Vision and Pattern Recognition (CVPR)}, New Orleans, US, 2022.

\bibitem{barron2021mip}
J.~T. Barron, B.~Mildenhall, M.~Tancik, P.~Hedman, R.~Martin-Brualla, and P.~P. Srinivasan, ``Mip-nerf: A multiscale representation for anti-aliasing neural radiance fields,'' in \emph{Proceedings of the IEEE/CVF International Conference on Computer Vision}, 2021, pp. 5855--5864.

\bibitem{barron2022mip}
J.~T. Barron, B.~Mildenhall, D.~Verbin, P.~P. Srinivasan, and P.~Hedman, ``Mip-nerf 360: Unbounded anti-aliased neural radiance fields,'' in \emph{Proceedings of the IEEE/CVF Conference on Computer Vision and Pattern Recognition}, 2022, pp. 5470--5479.

\bibitem{xiangli2022bungeenerf}
Y.~Xiangli, L.~Xu, X.~Pan, N.~Zhao, A.~Rao, C.~Theobalt, B.~Dai, and D.~Lin, ``Bungeenerf: Progressive neural radiance field for extreme multi-scale scene rendering,'' in \emph{European conference on computer vision}.\hskip 1em plus 0.5em minus 0.4em\relax Springer, 2022, pp. 106--122.

\bibitem{cui2024letsgo}
J.~Cui, J.~Cao, Y.~Zhong, L.~Wang, F.~Zhao, P.~Wang, Y.~Chen, Z.~He, L.~Xu, Y.~Shi \emph{et~al.}, ``Letsgo: Large-scale garage modeling and rendering via lidar-assisted gaussian primitives,'' \emph{arXiv preprint arXiv:2404.09748}, 2024.

\bibitem{zwicker2001ewa}
M.~Zwicker, H.~Pfister, J.~Van~Baar, and M.~Gross, ``Ewa volume splatting,'' in \emph{Proceedings Visualization, 2001. VIS'01.}\hskip 1em plus 0.5em minus 0.4em\relax IEEE, 2001, pp. 29--538.

\bibitem{schonberger2016structure}
J.~L. Schonberger and J.-M. Frahm, ``Structure-from-motion revisited,'' in \emph{Proceedings of the IEEE conference on computer vision and pattern recognition}, 2016, pp. 4104--4113.

\bibitem{hoppe2023progressive}
H.~Hoppe, ``Progressive meshes,'' in \emph{Seminal Graphics Papers: Pushing the Boundaries, Volume 2}, 2023, pp. 111--120.

\bibitem{park2021nerfies}
K.~Park, U.~Sinha, J.~T. Barron, S.~Bouaziz, D.~B. Goldman, S.~M. Seitz, and R.~Martin-Brualla, ``Nerfies: Deformable neural radiance fields,'' in \emph{Proceedings of the IEEE/CVF International Conference on Computer Vision}, 2021, pp. 5865--5874.

\bibitem{martin2021nerf}
R.~Martin-Brualla, N.~Radwan, M.~S. Sajjadi, J.~T. Barron, A.~Dosovitskiy, and D.~Duckworth, ``Nerf in the wild: Neural radiance fields for unconstrained photo collections,'' in \emph{Proceedings of the IEEE/CVF Conference on Computer Vision and Pattern Recognition}, 2021, pp. 7210--7219.

\bibitem{tancik2022block}
M.~Tancik, V.~Casser, X.~Yan, S.~Pradhan, B.~Mildenhall, P.~P. Srinivasan, J.~T. Barron, and H.~Kretzschmar, ``Block-nerf: Scalable large scene neural view synthesis,'' in \emph{Proceedings of the IEEE/CVF Conference on Computer Vision and Pattern Recognition}, 2022, pp. 8248--8258.

\bibitem{bojanowski2017optimizing}
P.~Bojanowski, A.~Joulin, D.~Lopez-Paz, and A.~Szlam, ``Optimizing the latent space of generative networks,'' \emph{arXiv preprint arXiv:1707.05776}, 2017.

\bibitem{knapitsch2017tanks}
A.~Knapitsch, J.~Park, Q.-Y. Zhou, and V.~Koltun, ``Tanks and temples: Benchmarking large-scale scene reconstruction,'' \emph{ACM Transactions on Graphics (ToG)}, vol.~36, no.~4, pp. 1--13, 2017.

\bibitem{hedman2018deep}
P.~Hedman, J.~Philip, T.~Price, J.-M. Frahm, G.~Drettakis, and G.~Brostow, ``Deep blending for free-viewpoint image-based rendering,'' \emph{ACM Transactions on Graphics (ToG)}, vol.~37, no.~6, pp. 1--15, 2018.

\bibitem{turki2022mega}
H.~Turki, D.~Ramanan, and M.~Satyanarayanan, ``Mega-nerf: Scalable construction of large-scale nerfs for virtual fly-throughs,'' in \emph{Proceedings of the IEEE/CVF Conference on Computer Vision and Pattern Recognition}, 2022, pp. 12\,922--12\,931.

\bibitem{lin2022capturing}
L.~Lin, Y.~Liu, Y.~Hu, X.~Yan, K.~Xie, and H.~Huang, ``Capturing, reconstructing, and simulating: the urbanscene3d dataset,'' in \emph{European Conference on Computer Vision}.\hskip 1em plus 0.5em minus 0.4em\relax Springer, 2022, pp. 93--109.

\bibitem{wang2004image}
Z.~Wang, A.~C. Bovik, H.~R. Sheikh, and E.~P. Simoncelli, ``Image quality assessment: from error visibility to structural similarity,'' \emph{IEEE transactions on image processing}, vol.~13, no.~4, pp. 600--612, 2004.

\bibitem{zhang2018unreasonable}
R.~Zhang, P.~Isola, A.~A. Efros, E.~Shechtman, and O.~Wang, ``The unreasonable effectiveness of deep features as a perceptual metric,'' in \emph{Proceedings of the IEEE conference on computer vision and pattern recognition}, 2018, pp. 586--595.

\bibitem{snavely2006photo}
N.~Snavely, S.~M. Seitz, and R.~Szeliski, ``Photo tourism: exploring photo collections in 3d,'' in \emph{ACM siggraph 2006 papers}, 2006, pp. 835--846.

\bibitem{jin2021image}
Y.~Jin, D.~Mishkin, A.~Mishchuk, J.~Matas, P.~Fua, K.~M. Yi, and E.~Trulls, ``Image matching across wide baselines: From paper to practice,'' \emph{International Journal of Computer Vision}, vol. 129, no.~2, pp. 517--547, 2021.

\end{thebibliography}

\end{document}